\documentclass[lettersize,journal]{IEEEtran}
\usepackage{color}
\usepackage{enumitem,xcolor}
\newlist{coloritemize}{itemize}{1}
\setlist[coloritemize]{label=\textcolor{blue}{\textbullet}}

\usepackage{cite}

\ifCLASSINFOpdf
   \usepackage[pdftex]{graphicx}
\else
\fi
\usepackage{slashbox}
\usepackage{makecell}
\usepackage{multirow}

\usepackage[cmex10]{amsmath}
\usepackage{amssymb,amsfonts}
\usepackage{siunitx} 

\ifCLASSOPTIONcompsoc
 \usepackage[caption=false,font=normalsize,labelfont=sf,textfont=sf]{subfig}
\else
 \usepackage[caption=false,font=footnotesize]{subfig}

 \usepackage{dblfloatfix}

\usepackage{url}

\usepackage{todonotes} 

\usepackage{soul}

\usepackage{bm}
\usepackage{nomencl}
\makenomenclature

\usepackage{arydshln}  

\DeclareMathAlphabet{\mathcal}{OMS}{cmsy}{m}{n}  
\DeclareMathAlphabet\mathbfcal{OMS}{cmsy}{b}{n} 

\def\Mimp{{$\mathbfcal{M}_{gr}$}} 
\def\Mf{{$\mathbfcal{M}_{pt}$}} 
\def\L{{$\mathcal{L}_1$}} 
\def\lgan{{$\mathcal{L}_{GAN}$}} 
\def\lcgan{{$\mathcal{L}_{cGAN}$}} 
\def\lsgan{{$\mathcal{L}_{LSGAN}$}} 
\def\lfft{{$\mathcal{L}_{FFT}$}} 
\def\ltot{{$\mathcal{L}_{tot}$}} 

\def\D{{\textit{D}}}
\def\G{{\textit{G}}}

\newcommand{\figr}{Fig.~\ref}				
\newcommand{\tabref}{Table~\ref}				
\newcommand{\Sec}{Section~\ref}					

\newcommand{\sena}{Sentinel-1A }
\newcommand{\senb}{Sentinel-1B }
\newcommand{\sen}{Sentinel-1 }

\newcommand{\hole}{Hole}
\newcommand{\tyr}{Tyristrand}
\newcommand{\nl}{Nordre Land}

\begin{document}
\title{Forest Parameter Prediction by Multiobjective Deep Learning of Regression Models Trained with Pseudo-Target Imputation}

\author{Sara~Björk,
        Stian~N.\ Anfinsen,
        Michael Kampffmeyer,
        Erik~N\ae sset,  
        Terje~Gobakken, 
        and Lennart Noordermeer
\thanks{Manuscript received ; revised .}
\thanks{S.\ Björk is with the Department of Physics and Technology, UiT The Arctic University of Norway, 9037 Troms\o, Norway and the Earth Observation Team, Kongsberg Satellite Services, 9011 Tromsø, Norway (e-mail: sara.bjork@ksat.no).}
\thanks{S.\ N.\ Anfinsen is with the Earth Observation Group, NORCE Norwegian Research Institute, 9019 Troms\o, Norway and the Department of Physics and Technology, UiT The Arctic University of Norway, 9037 Troms\o, Norway.}
\thanks{Michael Kampffmeyer is with the Department of Physics and Technology, UiT The Arctic University of Norway, 9037 Troms\o, Norway.}
\thanks{E.\ N\ae sset,  T.\ Gobakken and L.\ Noordermeer are with Faculty of Environmental Sciences and Natural Resource Management, Norwegian University of Life Sciences, 1432 \AA s, Norway.}
}

\maketitle

\begin{abstract}
In prediction of forest parameters with data from remote sensing (RS), regression models have traditionally been trained on a small sample of ground reference data. This paper proposes to impute this sample of true prediction targets with data from an existing RS-based prediction map that we consider as pseudo-targets. This substantially increases the amount of target training data and leverages the use of deep learning (DL) for semi-supervised regression modelling. We use prediction maps constructed from airborne laser scanning (ALS) data to provide accurate pseudo-targets and free data from Sentinel-1's C-band synthetic aperture radar (SAR) as regressors. A modified U-Net architecture is adapted with a selection of different training objectives. We demonstrate that when a judicious combination of loss functions is used, the semi-supervised imputation strategy produces results that surpass traditional ALS-based regression models, even though \sen data are considered as inferior for forest monitoring. These results are consistent for experiments on above-ground biomass prediction in Tanzania and stem volume prediction in Norway, representing a diversity in parameters and forest types that emphasises the robustness of the approach.
\end{abstract}


\IEEEpeerreviewmaketitle

\section{Introduction}
\label{sec:intro}
Accurate monitoring of forest above-ground biomass (AGB) is essential to better understand the carbon cycle. Vegetation biomass is, for example, a larger global storage of carbon than the atmosphere \cite{Kaasalainen2015, biomass}. Additionally, to monitor, measure and predict the amount of available AGB correctly is important for economic aspects, e.g.\ to estimate available raw materials or the potential for bioenergy \cite{johanssonBiomassProductionNorway1999, ericssonBioenergyPolicyMarket2004}. 

As the stem volume (SV) accounts for the highest proportion of biomass in each tree, typically 65-80\% \cite{seguraAllometricModelsTree2005, urbanAboveandBelowgroundBiomass2015, marklund1988biomass}, AGB monitoring often focuses on the available SV. In other applications, the total amount of available biomass is of interest, which comprises stems, stumps, branches, bark, seeds and foliage \cite{biomass, Galidaki,johanssonBiomassProductionNorway1999}.
Today, remote sensing (RS) data from radar, optical or airborne laser scanning systems (ALS) are commonly used together with a sparse sample of collected ground reference forest measurements to develop prediction models over larger areas and regions \cite{naessetMappingEstimatingForest2016, noordermeerComparingAccuraciesForest2019, solbergEstimatingSprucePine2010}. 

Satellite and airborne RS have become an important source of information about these forest parameters and others.
Traditionally, AGB and SV prediction models use relatively simple statistical regression algorithms, such as multiple linear regression, or machine learning regression models like random forests or multilayer perceptrons (MLPs) \cite{s.bjorkPotentialSequentialNonsequential2022}. These models are usually noncontextual, as they restrict the regressor information to the pixel that is being predicted and do not combine regressor and regressand information from neighbouring pixels, known as spatial context. 

Remote sensing is commonly used to infer forest parameters on spatial scales that are coarser than the pixel size, for instance on stand level. Hence, there is no formal reason to avoid the use of contextual information and one should select the method that provides the highest accuracy on the desired scale. This motivates the use of deep learning (DL) and convolutional neural networks (CNNs), whose popularity hinges on their efficient use of spatial context and the inference accuracy obtained by these highly flexible function approximators. The ability of CNNs to exploit spatial patterns was also pointed out in a recent review \cite{kattenborn2021review} as an explanation as to why CNN are particularly suitable for RS of vegetation.

A recent review \cite{hamedianfarDeepLearningForest2022} of DL methods applied to forestry concludes that these are in an early phase, although some work has emerged. We build our proposed method on Björk \emph{et al}'s sequential approach to forest biomass prediction \cite{s.bjorkPotentialSequentialNonsequential2022}, which uses a conditional generative adversarial network (cGAN) to generate AGB prediction maps by using synthetic aperture radar (SAR) as regressors and AGB predictions from ALS as the regressand. Their regression approach consists of two models that operate in sequence to provide more target data for training the model that regresses on SAR data. This implies that the first regression model learns the mapping between a small set of ground reference data and RS data from a sensor known to provide a high correlation with the response variable. ALS data are suitable for this purpose \cite{zolkos, Galidaki}, but are expensive to acquire. Hence, the second model in the sequence establishes a relationship between the ALS-derived prediction map, as a surrogate for the ground reference data, and RS data from a sensor that offers large data amounts at low cost, namely the Sentinel-1 SAR sensors.

This paper preserves some of the principal ideas from \cite{s.bjorkPotentialSequentialNonsequential2022}: The first is to train the regression model on an ALS-derived prediction map of the target forest parameter to increase the amount of training data. The motivation is that the small amount of ground reference data used to train conventional models limits their ability to capture the dynamics of the response variable, as demonstrated in \cite{s.bjorkPotentialSequentialNonsequential2022}. The second is to carry forward the use of CNNs to leverage their exploitation of contextual information, their flexibility as regression functions, and their demonstrated performance in other applications. 

At the same time, we make several new design choices to improve on the previous approach and remedy its weaknesses: Firstly, the sequential model is replaced by an approach where ground reference data are imputed with data from the ALS-derived prediction map. In practice, this is done by inserting the sparse set of true targets into the dense map of pseudo-targets. By letting these data sources together form the prediction target, the SAR-based prediction model can be trained simultaneously on ground reference data and the ALS-derived prediction map in a problem setting that we frame as semi-supervised learning; 
A second improvement is that we replace or combine the generative adversarial network (GAN) loss used in \cite{s.bjorkPotentialSequentialNonsequential2022} with a pixel-wise error loss and a frequency-aware spectral loss. This modification is motivated by an emerging awareness that the GAN loss used by the Pix2Pix \cite{isola} model employed in \cite{s.bjorkPotentialSequentialNonsequential2022} may be well suited to preserve perceptual quality and photo-realism, which is required in many image-to-image translation tasks, but is less appropriate for the regression task that we address. 

This paper has a stronger technical and methodological focus than \cite{s.bjorkPotentialSequentialNonsequential2022} and emphasises the method's ability to handle different tasks and cases: It demonstrates the proposed regression framework both on AGB prediction in dry tropical forests in Tanzania and on SV prediction in boreal forests in Norway, representing different parameters and very different forest types. Another difference is that the ALS-derived SV predictions used as pseudo-targets in the Norwegian dataset cover spatially non-contiguous forest stands, and is not a wall-to-wall prediction map. We have adapted the CNN-based regression algorithm for use with such data by implementing masked computation of the loss functions. 

In summary, we make the following contributions:
\begin{enumerate}
    \item We develop a method that enables us to train contextual deep learning models to predict forest parameters from C-band SAR data from the Sentinel-1 satellite. 
    \item We enable the CNN-based regression model to use target data that consist of spatially disjoint polygons, thereby showing that it can be trained on complex datasets that arise in operational forest inventories.
    \item By testing the method on AGB prediction in Tanzania and SV prediction in Norway, we demonstrate that it can handle different forest parameters and forest types.
    \item  We investigate an established consensus from the image super-resolution (SR) field about the trade-off between reconstruction accuracy and perceptual quality. For this purpose, we perform an ablation study of composite cost functions, including the GAN loss, a pixel-wise loss, and a recently proposed frequency loss.
    \item We demonstrate state-of-the-art prediction performance on datasets from Tanzania and Norway. Notably, we show that a deep learning model with C-band SAR data as input supercedes a conventional ALS-based prediction model after it has been trained on ground reference data imputed with ALS-derived predictions of the forest parameters.
\end{enumerate}

The remainder of this paper is organised as follows: \Sec{sec:related} reviews published research on related topics in deep learning applied to forest parameter prediction and other topics relevant to the proposed method. \Sec{sec:data} presents the datasets used in this work. \Sec{sec:method} details the proposed approach and describes how we facilitate the imputation of pseudo-targets for regression modelling, enabling the CNN model to learn from continuous and discontinuous target data using a variety of loss functions. Experimental results are provided in \Sec{sec:res} and discussed in \Sec{sec:disc}. Finally, \Sec{sec:conc} concludes the paper.

\section{Related work}\label{sec:related}
Björk \emph{et al.} showed in a precursor of this paper \cite{s.bjorkPotentialSequentialNonsequential2022} that the popular cGAN architecture \textit{Pix2pix} \cite{isola} can be used in the forestry sector to predict AGB from \sen data by training it on ALS-derived prediction maps. Their work inspired \cite{leonhardtProbabilisticBiomassEstimation2022} to also exploit ALS-derived AGB prediction maps and cGANs to predict AGB from multispectral and radar imagery and to quantify aleatoric and epistemic uncertainty. Despite apparent similarities, the current paper distinguishes itself from both \cite{s.bjorkPotentialSequentialNonsequential2022} and \cite{leonhardtProbabilisticBiomassEstimation2022} in many ways. The differences from \cite{s.bjorkPotentialSequentialNonsequential2022} are discussed in \Sec{sec:intro} when listing the contributions of the paper. Just like \cite{s.bjorkPotentialSequentialNonsequential2022},  Leonhardt \emph{et al.} \cite{leonhardtProbabilisticBiomassEstimation2022} train their regression network with adversarial learning through a cGAN architecture, but pretrain the generator with a mean square error (MSE) loss to find a proper initialisation. Notably, their final goal is not point prediction in the MSE sense or according to similar metrics, but to develop probabilistic methods for AGB prediction that quantify uncertainty.

Another example of deep learning applied to AGB prediction is Pascarella \emph{et al.} \cite{pascarellaReUseREgressiveUnet2023}, who show that a traditional U-Net \cite{ronnebergerUnetConvolutionalNetworks2015} trained with a pixel-wise error loss can be used as a regression model to predict AGB from image patches of optical Sentinel-2 data. Compared to \cite{pascarellaReUseREgressiveUnet2023}, we focus on utilising data from the Sentinel-1 radar sensor that, as opposed to the optical Sentinel-2 sensor, can acquire data both at night and under cloudy conditions and is therefore a more reliable source of data. 

Besides these examples, the literature on deep learning for regression modelling of forest parameters is sparse. This is also pointed out in the review of the use of CNNs in vegetation RS conducted by Kattenborn \emph{et al.} \cite{kattenborn2021review}. It found that only 9\% of the studies surveyed focused on regression modelling and only 8\% were specifically related to forestry and forest parameter retrieval, such as biomass prediction. 
A recently published review by Hamedianfar \emph{et al.} \cite{hamedianfarDeepLearningForest2022} attributes this literature gap to the challenge of acquiring the large amounts of target data needed to train accurate contextual CNN models for forest. This has been a main motivation for using pseudo-targets from existing prediction maps to train our SAR-based prediction models. For further inspiration, we have had to look to alternative topics in the literature.

Another image processing task that has inspired us to consider alternative loss functions and combinations of these is image super-resolution (SR). Single-image SR techniques are trained in a similar fashion as regression models: A full-resolution image is often used as the prediction target and a reduced resolution version of it as predictor data (see e.g.\ \cite{ESRGAN}), which renders the problem a prediction task that resembles the one in regression. Both the regression and the single-image SR task can be solved with generative models, but it is noteworthy that the literature identifies the SR task as an attempt to achieve two conflicting goals: It should produce images with high perceptual quality, meaning that they should appear natural and realistic. At the same time, it should reconstruct the underlying truth, that is, the high-resolution version of the input image, as closely as possible \cite{Blau2015, yangDeepLearningSingle2019, ESRGAN, Ledig, Soh2019SR}. 

The SR literature associates GAN losses and adversarial training with the perceptual quality criterion, as these enforce realistic fidelity and crispness in the generated image. This is achieved at the expense of accurate reconstruction in the MSE sense, since the generator module of the GAN effectively learns to hallucinate the kind of spatial pixel configurations that fools the discriminator module, but does not consider pixel-wise reconstruction. 
On the other hand, pixel-wise losses such as error measures based on the $L_1$ and $L_2$ norm naturally reduce the reconstruction error, but lead to a blurry appearance of the generated image that is not realistic \cite{Blau2015,ESRGAN}.

This has made us realise that although the Pix2Pix model has established itself as a preferred standard model in image-to-image translation, its GAN loss and adversarial learning approach may be better suited for generative tasks where the result must be visually credible. This is not a concern in the regression of biophysical parameters, where regression performance in terms of root mean square error (RMSE), mean absolute error (MAE) or similar metrics is used to evaluate and rank methods. When training such models, one should therefore consider other loss functions or composite loss functions that support the relevant aspects of the regression task. The SR literature exemplifies ways of combining different loss functions, both regarding which losses to select and how they should interact \cite{ESRGAN, Blau2015}. For instance, different losses can be used sequentially in pretraining and fine-tuning, or they can be used simultaneously as a composite loss.

Although perceptual quality is not of the essence for prediction maps of forest parameters, it may still be worth including loss functions that promote sharpness and visual information fidelity as part of a composite loss. One particular class of loss functions we find interesting to investigate is frequency-aware losses. Their aim is to preserve the high-frequency content of the image, which can e.g.\ be related to forest boundaries, structure and texture. These have not previously been utilised in forest applications, and to a limited extent in SR, but relevant work is found in the more general computer vision literature, where issues referred to as Fourier spectrum discrepancy, spectral inconsistency, frequency bias or spectral bias have gained a lot of attention \cite{chen2021ssd, Durall2020Watch, Chandrasegaran2021ACloserLook, khayatkhoeiSpatialFrequencyBias2022, s.bjorkSimplerBetterSpectral2022, Czolbe2020Watson, wangFrequencyDiscrepancyReal2021}. These terms relate to CNN-based generative models' lack of ability to capture the image distribution's high-frequency components, leading to blurriness and low perceptual quality. 

Some claim that the spectral bias is caused by the up-sampling method, e.g.\ transposed convolutions, used by the generator network \cite{Durall2020Watch, Chandrasegaran2021ACloserLook, wangFrequencyDiscrepancyReal2021}. Thus, changing the up-sampling method in the last layer of the generator network has been suggested \cite{Chandrasegaran2021ACloserLook}. However, Björk \emph{et al.} \cite{s.bjorkSimplerBetterSpectral2022} claim that changing the up-sampling procedure in the last layer from transposed convolution to e.g.\ nearest-neighbour interpolation followed by standard convolution gives ambiguous results. Chen \emph{et al.} \cite{chen2021ssd} argue that the down-sampling modules in the discriminator network of the GAN are the issue, resulting in a generator network that lacks an incentive from the discriminator to learn high-frequency information of the data. However, more recent work  \cite{khayatkhoeiSpatialFrequencyBias2022} proves that the frequency bias must be rooted in the GAN's generator and not the discriminator. Hence, there has been a focus on modifying the generative training objective by incorporating a spectral or frequency-aware loss with the traditional spatial loss during training \cite{Durall2020Watch, Czolbe2020Watson, s.bjorkSimplerBetterSpectral2022}. 

The observations and lessons from the precursor paper \cite{s.bjorkPotentialSequentialNonsequential2022} and from the literature on SR and generic generative models prompts us to investigate if model accuracy improves when we combine loss functions and whether pretraining of the model is enough or if we can increase model performance with a fine-tuning phase. Among the loss functions we combine is a newly proposed frequency-aware loss: the simple but promising FFT loss \cite{s.bjorkSimplerBetterSpectral2022}. It has been shown to perform better than other more complex frequency-aware losses \cite{Czolbe2020Watson, Durall2020Watch} on experiments where it was used to train a generative variational autoencoder (VAE) \cite{kingmaAutoEncodingVariationalBayes2014}. As the FFT loss has previously only been evaluated on VAEs with images from common benchmark datasets \cite{s.bjorkSimplerBetterSpectral2022}, we contribute with new insight into its behaviour when employed for other models and tasks.

\begin{figure}
    \begin{center}
        \includegraphics[height=5cm, width=0.8\columnwidth]{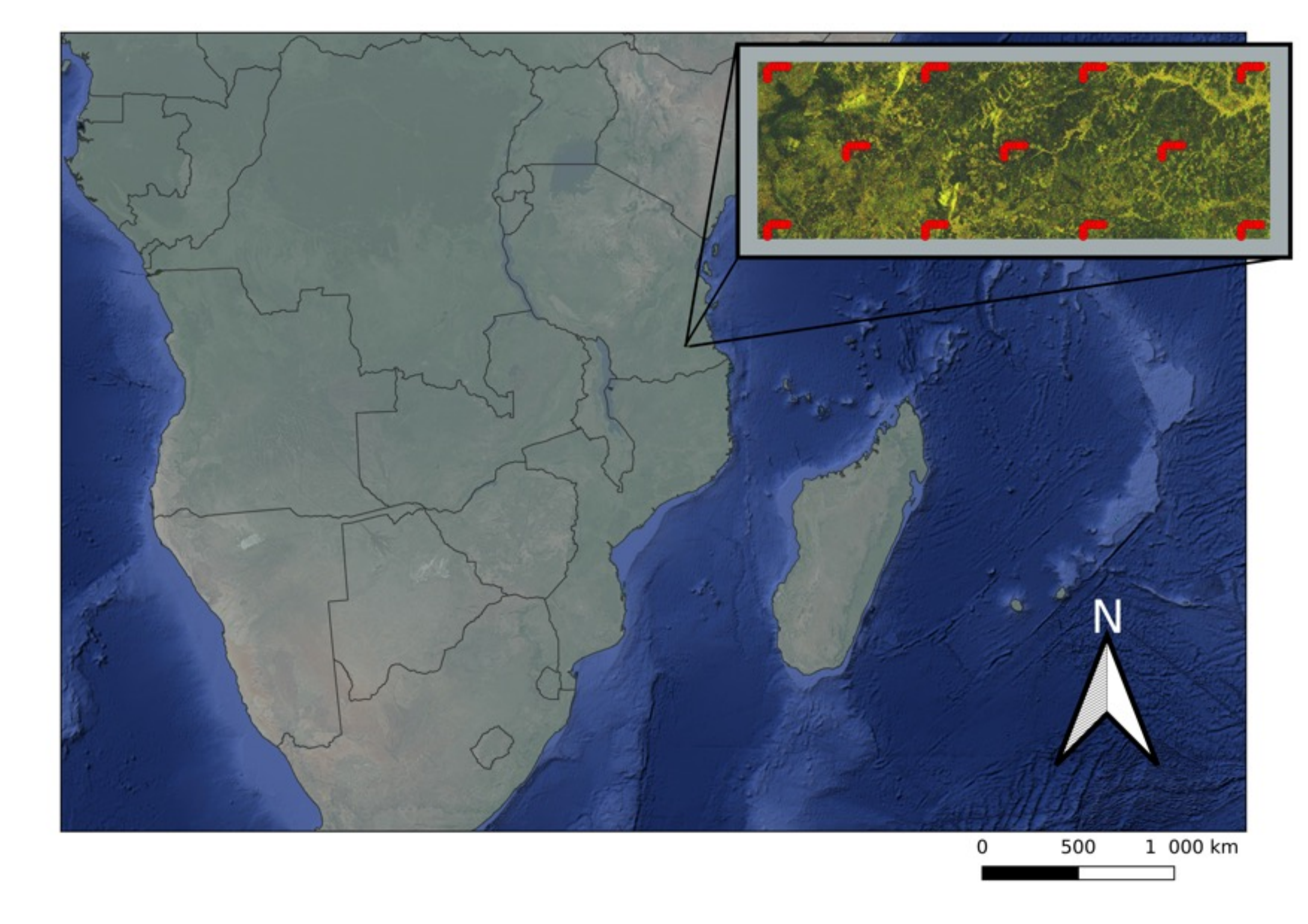}
    \caption{The location of the Tanzanian dataset, represented by \sena image data covering the AOI overlaid with ground reference data shown as red L-shaped clusters of ground plots. Figure from \cite{s.bjorkPotentialSequentialNonsequential2022}.} 
    \label{fig:mapT}        
    \end{center}
\end{figure}

\section{Study areas and datasets} \label{sec:data}
This section introduces the datasets used throughout this work, i.e.\ the ground reference target data, the ALS-derived prediction maps of AGB and SV, and the SAR data from the \sen sensors. The ALS-derived prediction maps will interchangeably be referred to as the pseudo-target datasets, while the ground reference data are also referred to as field data, data from the field plots, or true prediction targets.
The AGB dataset comes from the Liwale district in Tanzania. The SV datasets are from three regions in the southeast of Norway: Nordre Land, Tyristrand and Hole. 

For Tanzania, both the field data and the ALS data were acquired in 2014, as described in \cite{naessetMappingEstimatingForest2016} and \Sec{sec:tanzania}. The \sena satellite was launched in April 2014 and only one single \sena scene acquired in September 2015 was found to comply with our requirements, meaning that it covers one of Liwale's two yearly dry seasons and is close enough in time to the field inventory and the ALS campaigns in Tanzania. For Norway, the acquisition of the ALS data in 2016 and the field inventory in 2017 (see \cite{noordermeerComparingAccuraciesForest2019} and \Sec{sec:norway}) implies that more \sen data are available. Thus, the models we develop for the Norwegian test sites utilise a temporal stack of \sena and \senb scenes from July 2017. 

\subsection{Study area and dataset description}
\label{sec:dataset}
\subsection{Study area and dataset description}

This section briefly describes the Tanzanian and Norwegian study areas, including the ground reference data and related ALS-derived prediction maps. The interested reader is referred to \cite{naessetMappingEstimatingForest2016} and \cite{noordermeerComparingAccuraciesForest2019}, respectively, for in-depth descriptions of the ground reference data and the ALS-derived prediction maps. 

\subsubsection{Tanzanian study area}\label{sec:tanzania}
This work focuses on the same study area as \cite{s.bjorkPotentialSequentialNonsequential2022}, i.e.\ the Liwale district in the southeast of Tanzania (\ang{9}52'-\ang{9}58'S, \ang{38}19'-\ang{38}36'E). The area of interest (AOI) is a rectangular region with a size of $11.25 \times 32.50\ \mathrm{km}$ (WGS 84/UTM zone 36S). \figr{fig:mapT} shows the location of the AOI in Tanzania and the distribution of the 88 associated field plots. These field plots were collected within 11 L-shaped clusters, each containing eight plots, as seen in \figr{fig:mapT}. 

The field work was performed in January-February 2014, and a circular area of size 707 $\mathrm{m}^2$ represents each sample plot on the ground, i.e.\ they have a radius of 15 $\mathrm{m}$. We refer to \cite{Tomppo2014ASampling} for a description of the national level sample design in Tanzania, while \cite{ene2017, ene2016, naessetMappingEstimatingForest2016} explain how data from the field work are used to develop large-scale AGB models. Generally, the miombo woodlands of the AOI are characterised by a large diversity of tree species. Measured AGB from the field work ranged from 0 to 213.4 $\mathrm{Mg\,ha}^{-1}$ \cite{naessetMappingEstimatingForest2016} with a mean and standard deviation of $\mu\!=\!51.3$ and $\sigma=45.6\mathrm{Mg\,ha}^{-1}$. 

\begin{figure}
    \begin{center}
    \includegraphics[scale=0.5]{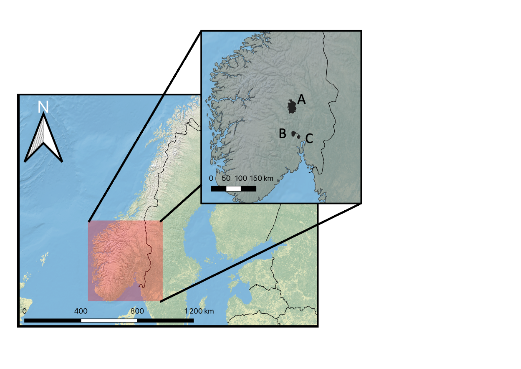}
    \caption{Location of the regions Nordre Land (A), Tyristrand (B) and Hole (C) in the Norwegian dataset.} 
    \label{fig:mapN}
     \end{center}
\end{figure}

\subsubsection{Tanzanian ALS-predicted AGB data} \label{sec:alsagb}
We follow \cite{s.bjorkPotentialSequentialNonsequential2022} and use the same ALS data from the Liwale AOI, which was acquired in 2014. We refer to \cite{naessetMappingEstimatingForest2016} for details on the ALS flight campaign, ALS data processing, and the match-up of ALS data with ground reference AGB data from the field plots. After model fitting, the ALS-based AGB model was in \cite{naessetMappingEstimatingForest2016} used to infer a wall-to-wall prediction map for the whole AOI in Liwale. The wall-to-wall map is represented as a grid with square pixels of size 707 $\mathrm{m}^2$. We have gained access to this prediction map and will use it as pseudo-targets to train contextual CNN models for AGB predictions based on \sen SAR data.

\subsubsection{Norwegian study area}\label{sec:norway}
The Norwegian study area consists of three regions shown in \figr{fig:mapN} and referred to as \nl\ (A), \tyr\ (B) and \hole\ (C). All field work was performed during the summer and fall of 2017, initially resulting in 386 circular field plots of shape 250 $\mathrm{m}^2$ distributed over the three regions. We refer to \cite{noordermeerComparingAccuraciesForest2019} for a description of the sampling design and related data properties. 

Of the original 386 field plots used for modelling stem volume, a total of 122 plots were not located within polygons of forest stands delineated in the inventories, and thus fell outside the spatial extent of the ALS-predicted SV datasets. We therefore excluded these plots from the analysis. In \tabref{tab:regions}, the column \textit{No. of plots (after filtering)} indicates the number of field plots included in the current study. The remaining entities of \tabref{tab:regions}, such as geographical coordinates, inventory size, field inventory information and distribution of the dominant tree species in each region, are sourced from \cite{noordermeerComparingAccuraciesForest2019}.

In \nl, ground reference values of SV ranged from 33.7 to 659.2 $\mathrm{m^3\,ha}^{-1}$ with a mean and standard deviation of $\mu\!=\!252.7$ and $\sigma\!=\!145.5\mathrm{m^3\,ha}^{-1}$. In \tyr\ it ranged from 56.1 to 513.3 $\mathrm{m^3\,ha}^{-1}$ with $\mu=212.6$ and $\sigma=96.9\mathrm{m^3\,ha}^{-1}$, while in \hole\ it ranged from 29.5 to 563.9 $\mathrm{m^3\,ha}^{-1}$ with $\mu=253.4$ and $\sigma=125.8\mathrm{m^3\,ha}^{-1}$.

\begin{table*}[]
\caption{Characteristics of each of the three Norwegian regions included in this work. All entities, except for column \textit{No. of plots (after filtering)}, which refers to the field plots that are used for this work, are borrowed from \cite{noordermeerComparingAccuraciesForest2019}.} 
    \label{tab:regions}
    \centering
    \begin{tabular}{c c l c c c c c c}
         \hline
  Region & Name & Geographical & Inventory &  No.\ of plots &  Proportion & Proportion & Proportion\\
 & & coordinates &  size ($\mathrm{km}^2$) &  (after filtering) & spruce & pine & deciduous\\
   \hline
  A & Nordre Land & 60$^{\circ}$50´N,& 490      & 136   & 74\%  & 23\%  & 3\% \\
   & & 10$^{\circ}$85´E &  &  &    &   &    &  \\
  B & Tyristrand & 60$^{\circ}$6´N, & 60      & 77   & 15\%  & 80\%  & 5\% \\
  &  & 10$^{\circ}$20´E &  &   &  &    &   &  \\  
  C & Hole   & 60$^{\circ}$1´N,  & 45        & 51 & 89\%  & 4\%  & 7\% \\
   &  & 10$^{\circ}$20´E &  &   &   &  &   &  \\
    \hline 
    \end{tabular}
\end{table*}

\subsubsection{Norwegian ALS-predicted SV data} \label{sec:alsvol}
The ALS flight campaigns were performed in 2016 for all three regions of Norway. We refer to \cite{noordermeerComparingAccuraciesForest2019} for a description of how the ALS data were processed, the formulation of the nonlinear local prediction models and the match-up of ALS-derived predictions with ground reference data. After model fitting, maps of SV predictions were generated for all three regions, limited to areas where the forest height exceeded 8-9 meters. We refer to these as the ALS-derived SV prediction maps. In all regions, predictions were made for square pixels of size 250 $\mathrm{m}^2$, i.e. $15.8\ \mathrm{m} \times 15.8\ \mathrm{m}$ on the ground. The ALS-derived SV is given in units of $\mathrm{m^3\,ha}^{-1}$. 

\begin{figure}
\begin{center}
    \includegraphics[height=7cm, width=0.8\columnwidth]{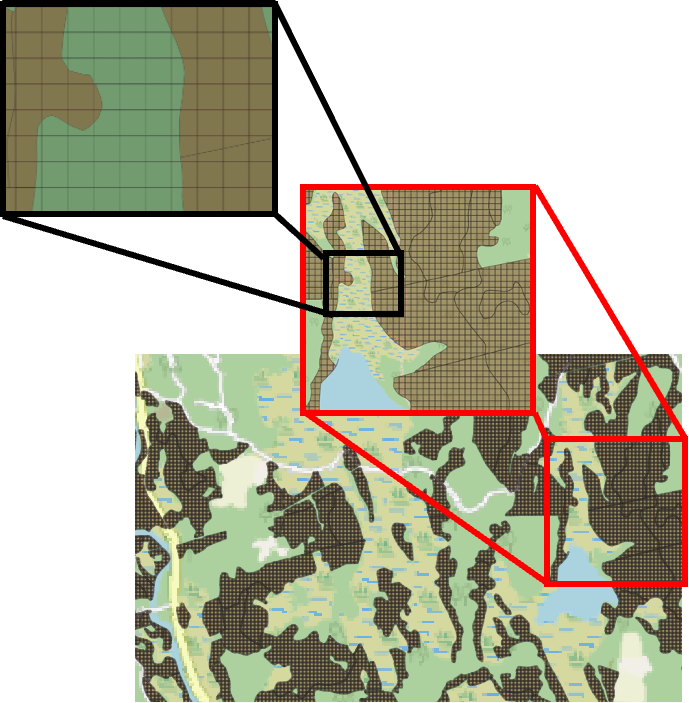}
    \caption{Small section of the ALS-derived SV prediction map from Nordre Land. SV has been predicted in the brown areas. The lattice represents the common pixel grid of the SAR predictor data and the rasterised SV prediction map. The original prediction map is obtained in vector format, with one SV prediction per polygon and multiple polygons per grid cell. The rasterisation process with merging of polygons is described in detail in the text.} 
    \label{fig:discNL}
    \end{center}
\end{figure}

\subsection{Postprocessing of the ALS-derived prediction maps} \label{sec:postals}
The ALS-derived prediction maps have been obtained as vector data in polygon format stored as shapefiles. These must be converted to raster data in order to be used as training data for CNN models. This conversion is straightforward for the Tanzania datasets, where all polygons are square and have the same areal coverage. Hence, we map project and sample the SAR data such that the SAR pixels coincide with the polygons of the AGB prediction map. 

The process for the Norway dataset is more complicated. \figr{fig:discNL} shows a section of the ALS-derived SV prediction map retrieved in the Nordre Land municipality. Brown areas show where SV predictions are available, whereas the background (other colours) is retrieved from OpenStreetMap \cite{openstreetmapcontributorsPlanetDumpRetrieved2017}. 
An overlaid lattice of square grid cells can be seen at all zoom levels of \figr{fig:discNL}. This lattice represents two things: Firstly, it contributes to the delineation of the polygons in the SV prediction map. In this dataset, SV has been predicted for polygons of varying size and shape, that are delimited by: 1) the grid cells of the lattice, as mentioned above; 2) the commercial forest boundaries that enclose the brown areas; and 3) curves within the brown areas that mark internal forest boundaries and subdivide different forest areas. These are seen at all zoom levels of the figure. Secondly, the lattice coincides with the map grid of the SAR data, since we have map projected and resampled the SAR images to align their map grid with the lattice of the SV polygons. Hence, the lattice grid is identical to the pixel grid we want for our training dataset.

In summary, SV is only predicted in brown areas. Each prediction is associated with a polygon, which can be square if it is only delimited by the lattice and coincides with a lattice grid cell. It can also be of irregular shape and size, if a forest perimeter or an internal forest boundary delimits it. Each polygon is assigned a stem volume, $V$, and an areal coverage, $A$. Some of the square lattice cells are fully covered by one or more polygons, while others are only partly covered. Some lattice cells contain one polygon, while others contain two or more.
We refer to this as a multipolygon format, as every lattice grid cell potentially contains multiple polygons. 

The multipolygon dataset must be rasterised into a target dataset with the same pixel grid as the SAR predictor data. This means that all polygons within a lattice grid cell must be merged, and the grid cell must be assigned a single SV value and the associated areal coverage. The predicted SV contributed by all intersecting multipolygons is computed as

\begin{equation}
 V_{merged} = \sum_{i=1}^n{V_{mp(i)}},
\end{equation}
where \textit{mp(i)} indicates multipolygon number $i$ and $n$ is the number of multipolygons in a grid cell. 
Simultaneously, the total areal coverage is computed as: 
\begin{equation}
A_{merged} = \sum_{i=1}^n{A_{mp(i)}}.
\end{equation}

The described merging process guarantees that each grid cell is assigned a unique SV, but this value does not necessarily represent a full grid cell of 250 $\mathrm{m}^2$. To quality assure the SV dataset, we remove all SV predictions with less than 40\% areal grid cell coverage. This threshold is chosen heuristically to accommodate all three regions, as this removes less than 12\% of the \nl\ and \tyr\ dataset and less than 10\% of the \hole\ dataset. The remaining SV prediction dataset is deemed suitable for the training of CNN regression models. All postprocessing steps are applied using QGIS \cite{qgis}.
\subsection{SAR data}
Low data cost can sometimes be crucial for developing forest parameter monitoring systems suitable for commercial or operational use. This paper utilises SAR data from the freely available \sen sensors, which also offer short revisit time and good coverage for the areas of interest. The SAR images are dual-polarisation (VV and VH) C-band scenes acquired in a high-resolution Level-1 ground range detected (GRD) format with a 10 $\mathrm{m}$ pixel size. The SAR data was downloaded from Copernicus Sentinel Scientific Data Hub\footnote{See \url{https://scihub.copernicus.eu/dhus/#/home}}.

For the AOI in Tanzania, we use a single scene acquired on 15 September 2015, as this is the only available \sen product that covers the AOI at a time close to the acquisition of the ALS data and during one of Liwale's two yearly dry seasons. The latter criterion implies that the radar signal achieves sufficient sensitivity to dynamic AGB levels. 

We utilise data from the Sentinel-1A and -1B satellites for the three Norwegian regions. Since the field work for the three Norwegian regions was performed during the summer and fall of 2017, we decided to create temporal stacks of \sen-scenes from July 2017 for each of the three regions. 

\subsection{SAR data processing and preparation of datasets}
The Sentinel-1 GRD product in the Tanzanian dataset was processed with the ESA SNAP toolbox \cite{snap} following the workflow described in \cite{s.bjorkPotentialSequentialNonsequential2022}.

The \sen GRD products in the Norway dataset have been processed with the GDAR SAR processing software at NORCE Norwegian Research Institute. They are geocoded with a $10\ \mathrm{m}\times 10\ \mathrm{m}$ digital elevation model to the same map projection as the ALS-derived SV prediction map and resampled to a pixel resolution of $15.8\ \mathrm{m}$ to match the $250\ \mathrm{m^2}$ grid cells of the prediction map.
Since \cite{s.bjorkPotentialSequentialNonsequential2022} showed that it is more advantageous to train CNN-based prediction models with \sen intensity data on decibel (dB) scale, the stacks of \sen scenes for the Norwegian regions are converted to dB format. The final \sen products for the Norwegian regions contain nine features that were extracted from the \sen time series: \textrm{NDI}, \textrm{mean}(VV), \textrm{mean}(VH), \textrm{min}(VV), \textrm{min}(VH), \textrm{max}(VV), \textrm{max}(VH), \textrm{median}(VV), \textrm{median}(VV). NDI denotes the normalised difference index feature, a normalised measure of how much the measured backscatter differs in VV and VH. It is computed as

\begin{equation}
    NDI = (VV-VH)/(VV+VH).
    \label{eq:ndi}
\end{equation}

\section{Methodology}
\label{sec:method}
This section describes the proposed methodology to train contextual CNN models for forest parameter prediction. We describe the semi-supervised approach and how training, test and validation datasets are created for each region. In general terms, we introduce the CNN models we use in our work and describe the changes proposed to improve on the performance obtained in \cite{s.bjorkPotentialSequentialNonsequential2022}. This section focuses on a semi-supervised learning strategy where we impute the sparse reference data with data from ALS-derived prediction maps to increase the amount of training data and to create a dataset that allows us to train CNN models. It also explains the multiobjective training approach, which exploits composite loss functions with varying objectives in the pretraining and fine-tuning stages.

\subsubsection{Overview} \label{sec:overview}
\begin{figure}
    \centering
    \includegraphics[height=8cm, width=0.8\columnwidth]{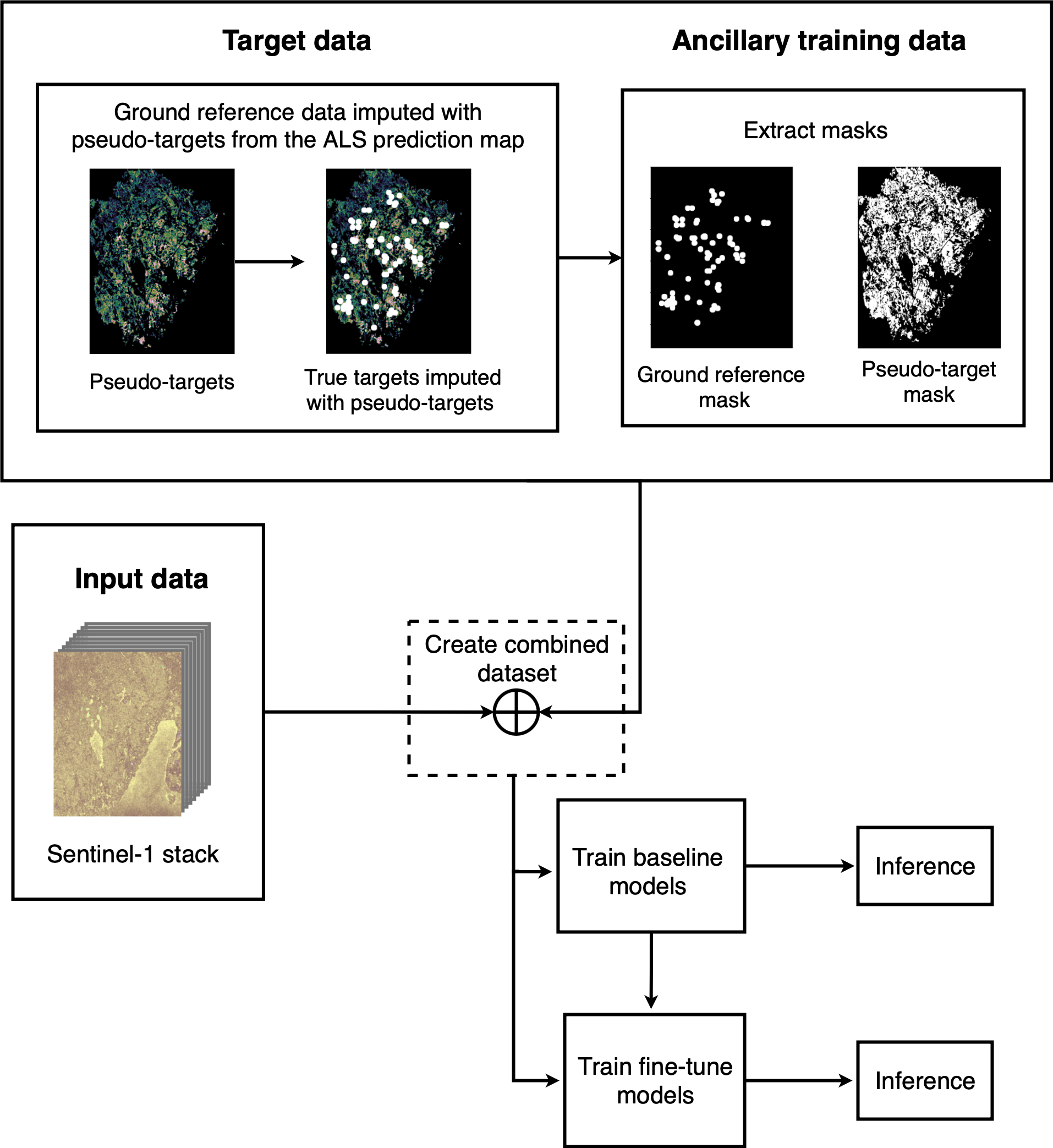}
    \caption{Overall workflow for dataset generation, model training and inference to create prediction maps, displayed with image data from the \tyr\ dataset. True targets (white circles) have been magnified for illustrative purposes.} 
    \label{fig:pipeline}
\end{figure}

The framework of the proposed method is illustrated in \figr{fig:pipeline}. Initially, the ground reference data, also known as the true prediction targets, are imputed with the ALS-derived prediction map, also called the pseudo-target dataset. Then two binary masks are created, one indicating the pixel positions of the true targets and the other indicating pixels where pseudo-target data are available. The two masks are referred to as ancillary training data. They enable the CNN to learn from discontinuous pseudo-target data and boost learning in regions where ground reference data are available. When the pseudo-target data are spatially continuous and have the same extent as the predictor data, the pseudo-target mask will have a constant value of one. The imputed target dataset and the two masks are combined with regressor data from the \sen sensor. See \Sec{sec:trtestval} and \figr{fig:patches} for details. \figr{fig:pipeline} shows that baseline models are pretrained as an initial training step. Following the pretraining stage, fine-tuning may be applied to the baseline CNN models with a composition of different losses. Inference, i.e.\ production of SAR-derived prediction maps, is done with the resulting models\footnote{Code will be available from\\
\url{https://github.com/sbj028/DeepConvolutionalForestParameterRegression}}. 

\subsection{Imputing ground reference data with pseudo-targets} \label{sec:pseudometh}
The cGAN-based models developed in \cite{s.bjorkPotentialSequentialNonsequential2022} for SAR-based regression trained on ALS-derived prediction maps could not compete with the conventional ALS-based regression model in terms of prediction accuracy. We argue that this is because the cGAN model is not trained on the true prediction targets and therefore inherits too much of the uncertainty in the ALS-based prediction maps. By contrast, the conventional ALS model was fitted directly to all the true prediction targets. To address this shortcoming and improve the performance of CNN models, we propose to impute pseudo-targets from the ALS-derived prediction maps into the dataset of true prediction targets, so that the CNN model is trained on the complete set of available targets. Since the ground reference dataset is much smaller than the prediction maps, this is in practice done by inserting true targets into the pseudo-target prediction maps. Following the imputation process, the Tanzanian dataset comprises less than 0.08\% of target values originating from the ground reference data. For the Norwegian datasets, the ground reference data represents less than 0.04\%, 0.11\%, and 0.13\% of the pixels in the respective \nl, \tyr, and Hole datasets after the imputation process.

We would generally use all available ground reference data for model training and hyperparameter tuning. However, for model evaluation, we report the performance after cross-validation (CV), where we have trained models on a target dataset that only contains 80\% of the true target labels. The remaining 20\% are reserved for validation. Results obtained with CV are referred to as CV-RMSE in the result section. 

\subsection{Preparing the datasets for contextual learning}\label{sec:trtestval}
\begin{figure*}
    \includegraphics[scale=0.15]{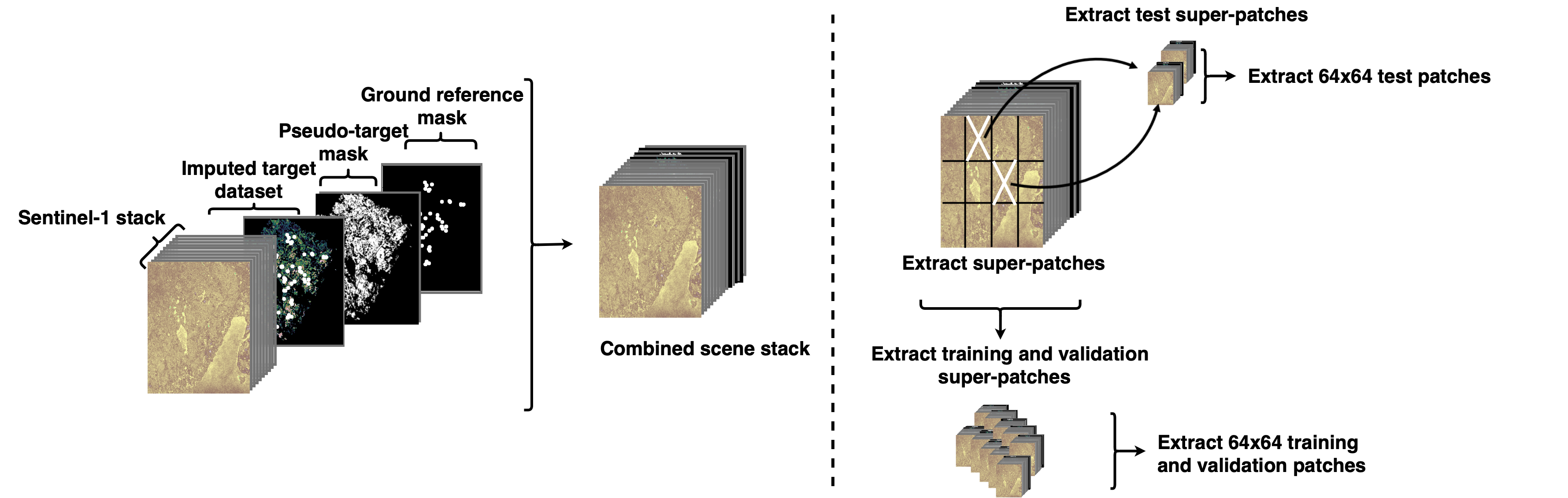}
    \caption{Illustration of how training and test image patches are extracted from the stack of \sen dataset, imputed target dataset, pseudo-target mask and ground reference mask. The datasets shown are retrieved from the AOI in \tyr. However, the process is identical for all Norwegian regions and representative of how the Tanzanian dataset is prepared. See \Sec{sec:trtestval} for details.} 
    \label{fig:patches}
\end{figure*}
 
To create training, test and validation datasets for the Norwegian regions, all true target labels from the field inventory were first inserted into the ALS prediction maps of pseudo-targets. Two binary masks were additionally created; the pseudo-target mask, denoted \Mf\, indicates the positions of available ALS-derived predictions. It is needed for masked computation of the loss functions, which are restricted to pixels where prediction targets are available. The ground reference mask, denoted \Mimp, holds the positions of the true prediction targets. It is also used in the loss computation, where we weight the loss for the true prediction targets higher than the pseudo-targets.

After having produced the imputed target dataset and the two masks, we follow the workflow shown in \figr{fig:patches} to create datasets with training, test and validation image patches. The figure illustrates the process for \tyr, but it is identical for all three Norwegian regions. Firstly, all available data are combined into a stack, including the \sen mosaic of nine feature bands, the imputed target map, and the two masks. Then the entire scene is divided into superpatches by splitting it into blocks with no overlap. A superpatch is defined as a block of pixels that is larger than the image patches we use for training, testing and validation. See \tabref{tab:patches} for an overview of the total number of pixels in each region, the corresponding size of each superpatch and the number of possible superpatches that can be extracted for that region. \Mf\ was used to remove superpatches with no overlap with pseudo-targets. Among all available superpatches, those with at least 10\% overlap with \Mf\ were identified as candidates for the test dataset. Fulfilling this criterion, approximately 15\% of all available superpatches were randomly selected as test superpatches. These were further split into test patches of $64\times 64$ pixels without overlap. Test patches having no overlap with \Mf\ were discarded. \tabref{tab:patches} shows each region's final number of test patches. 

The remaining superpatches were initially used for hyperparameter tuning. See Appendix \ref{app:hyp} for details. After this, all patches not used for testing were combined into training sets for the Norwegian models by splitting superpatches into training image patches of $64\times 64$ pixels using 50\% overlap and data augmentation with flipping and rotation. Patches with no overlap with \Mf\ were discarded. \tabref{tab:patches} lists the number of training image patches per region after hyperparameter tuning.

The training, test and validation datasets for Tanzania were created by similar use of superpatches. Since the Tanzanian ALS-derived prediction map covers the whole AOI without any discontinuities, there is no need to check if image patches overlap with pseudo-targets. See \tabref{tab:patches} for details on region sizes in pixels and the number of test and training patches. 

To evaluate the models, we also created CV target datasets where we used only 80\% of the true target labels from the field inventory. We compute the model's performance both when it is trained with all true target labels and also a CV performance for the case when 20\% of the true target labels are held out and used for testing. When comparing these results, one must recall that in the former case, the model has seen the test data during training. Moreover, the models are in the CV case trained with less true prediction targets. 

\begin{table}[]
\caption{Description of the different regions, information related to creating training and test patches and the final number of test and training patches per region. The listed number of training patches are after data augmentation. See \Sec{sec:trtestval} for details}
    \label{tab:patches}
    \centering
    \resizebox{\columnwidth}{!}{%
    \begin{tabular}{l c c c c c}
         \hline
                & Region & Superpatch & Number of     & Number of     & Number of \\
     & shape  & size        & superpatches & test patches  & training patches \\
               Region Name &  (pixels)           &             &               & ($64\times64$)  & ($64\times64$)  \\
   \hline
  Nordre Land & $3136\times1984$ & $224\times248$  & 128 & 87 & 18072 \\
  Tyristrand & $1152\times896$  & $128\times128$  & 63 & 25 & 2776 \\
 Hole   & $768\times768$  & $128\times128$ & 12 & 14 & 1384 \\
\hdashline
 Liwale & $423 \times 1222$  & -- & -- & 14 & 2784 \\
    \hline 
    \end{tabular}}
\end{table} 

\subsection{Backbone U-Net Implementation}\label{sec:backbone}
The CNN we use for SAR-based prediction of forest parameters is a modified version of the U-Net architecture in \cite{ronnebergerUnetConvolutionalNetworks2015}, a fully convolutional encoder-decoder network originally developed for biomedical image segmentation. The U-Net consists of a contraction part and a symmetric extraction path, with skip connections between each encoder block and the associated decoder block. The skip connections imply that low-level feature maps from the contraction part are concatenated with high-level feature maps from the extraction part to improve the learning in each level of the network. 

\begin{figure*}
    \centering
    \includegraphics[height=9cm, width=\textwidth]{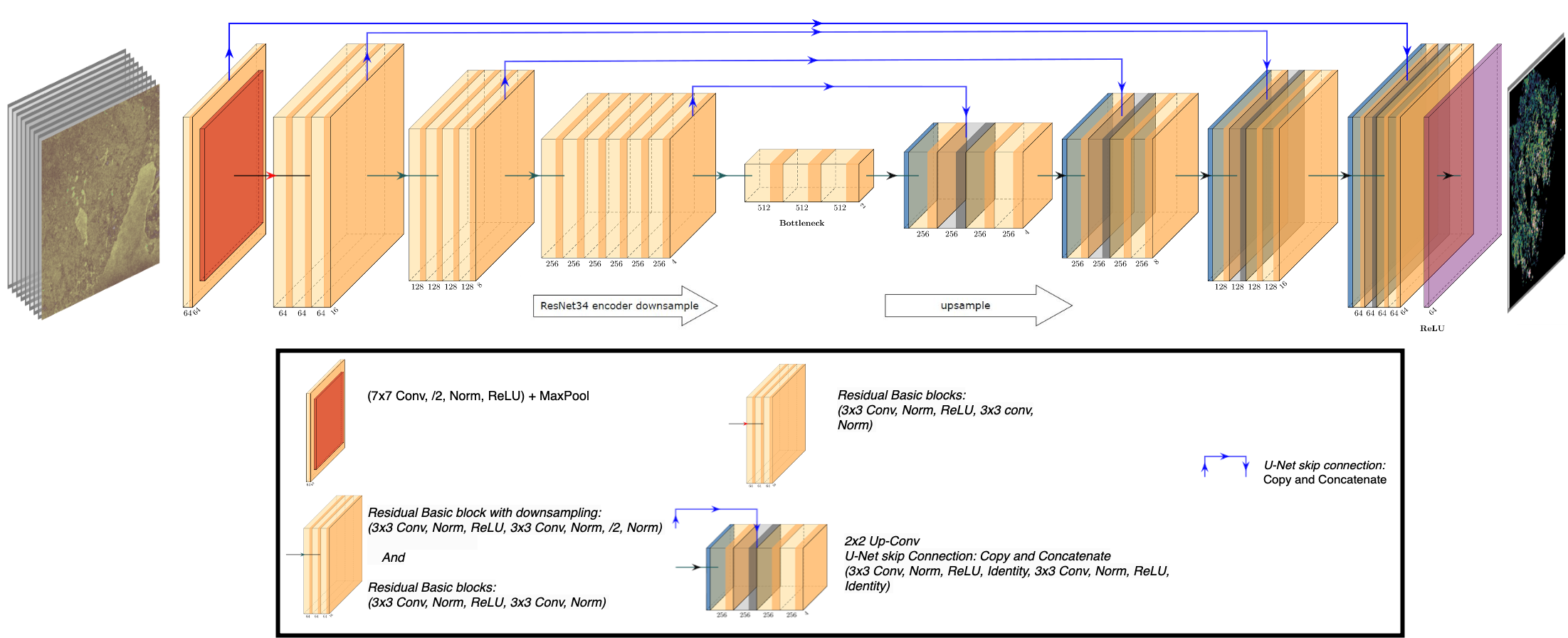}
    \caption{Above: Regression U-Net architecture used for image-to-image translation. Below: modules of the regression U-Net. Modification of figure in\cite{iqbalHarisIqbal88PlotNeuralNetV12018}.} 
    \label{fig:unet}
\end{figure*}

\figr{fig:unet} illustrates the U-Net generator network we use with an encoder-decoder depth of 4. This is the depth used by the Norwegian models, determined by hyperparameter tuning, while the Tanzanian models use a depth of 5. In both cases, we use ResNet34 \cite{resnet} as backbone for the convolutional encoder network and refer to the whole model as a regression U-Net.

The regression U-Net is trained to perform image-to-image translation. I.e., given \sen image patches from the input domain, the model translates these into prediction maps of AGB or SV maps for the same area, guided by the imputed target data.  
For the Norwegian datasets, we have modified the first layer of the encoder to enable nine-channel inputs, i.e.\ input tensors of dimension $9 \times 64 \times 64$. The Tanzanian models take three-channel inputs with a shape of $3 \times 64 \times 64$. Additionally, the segmentation head was removed from the original U-Net architecture, as our work concerns a regression task and not a segmentation task. Finally, the softmax activation function in the final layer was replaced with a ReLU activation function to ensure non-negative AGB and SV predictions. 

The initial layer of the encoder network uses a $7 \times 7$ convolution kernel with a stride of 2, followed by a normalisation layer, ReLU activation and a max-pooling operation. This implies that the number of feature channels is increased to 64, while the image dimension is decreased to $16 \times 16$ pixels. The following layer combines residual basic blocks, each using a $3 \times 3$ convolution, followed by a normalisation layer, ReLU activation, $3 \times 3$ convolution and a normalisation layer. The following encoder layers' residual blocks additionally employ down-sampling layers, which double the feature channels and half the spatial resolution of the image. In addition to the skip connections previously mentioned, each residual block uses common short connections \cite{resnet}.

Each block in the extraction part uses upsampling through nearest-neighbour interpolation and combines feature maps from the skip connection. It further processes the feature maps through two identical transformations, each including $3 \times 3$ convolutional filtering followed by a normalisation layer, ReLU activation and identity mapping. The upsampling procedure halves the number of feature maps while doubling the spatial resolution. We use the Pytorch implementation of the U-Net model from \cite{iakubovskiiSegmentationModelsPytorch2019} for our regression U-Net, with the above-mentioned modifications. 
\subsection{Pretraining Stage}
\label{sec:pretrain}
We follow the training procedure proposed for ESRGAN, a super-resolution model trained with multiple objectives in \cite{ESRGAN}, and divide the training of the U-Net architecture into two stages: pretraining and fine-tuning. In the pretraining stage, we train two baseline CNN models: an \L-based regression U-Net and a cGAN-based generative U-Net. In the fine-tuning stage, described in Section \ref{sec:finetune}, we continue to train the baseline models with additional losses.

\subsubsection{Pixel-aware Regression U-Net}\label{sec:regunet}
We refer to a regression-type U-Net model optimised on a pixel-wise loss computed between model-inferred predictions and target predictions as a pixel-aware regression U-Net (PAR U-Net). In this work, the PAR U-Net is optimised on the \L\ loss similar to \cite{ESRGAN}, i.e.\ 
\begin{equation}
    \mathcal{L}_1 = \sum_{k}{||\boldsymbol{Y} - \textit{F}(\boldsymbol{X})||} = \sum_{k}{||\boldsymbol{Y} - \boldsymbol{\hat{Y}||}},
    \label{eq:l1}
\end{equation}
where $\boldsymbol{X}$ and $\boldsymbol{Y}$ represent a corresponding pair of input and target image patches from the training dataset, $\boldsymbol{\hat{Y}}=\textit{F}(\boldsymbol{X})$ is the image patch predicted by a CNN model $\textit{F}(\cdot)$, and \textit{k} is the total number of image patches.

\subsubsection{cGAN U-Net}\label{sec:cgan}
In addition to training the modified U-Net with a \L\ loss, we also train it as a cGAN, like in \cite{s.bjorkPotentialSequentialNonsequential2022}. Formally conditioned on image patches from the \sen input domain, the cGANs generator (\G) network is trained to learn the optimal mapping $\G : \mathcal{X} \to \mathcal{Y}$ to generate realistic-looking image patches from the target domain. The \G\ network also uses the regression U-Net architecture in \figr{fig:unet}. 

Simultaneously as the \G\ network aims to improve the generation task, the adversarially trained discriminator network (\D) is trained to distinguish between real or false pairs of image patches successfully. A real pair of image patches corresponds to one \sen image patch and the corresponding target ALS-derived prediction map. On the other hand, a false pair corresponds to one \sen image patch and the corresponding prediction map generated by \G. Adversarial training of \G\ and \D\ results from optimising the minimax loss function of the so-called Vanilla GAN (VGAN) \cite{gan2014}: 
\begin{equation}
\begin{split}
\min_{\G} \max_{\D} \mathcal{L}_{VGAN}(\D,\G)   = \mathbb{E}_{\boldsymbol{X},\boldsymbol{Y}}[\text{log}\ \D(\boldsymbol{X},\boldsymbol{Y})] \\
 + \mathbb{E}_{\boldsymbol{X}}[\text{log}(1 - \D(\boldsymbol{X},\G(\boldsymbol{X})))].&
\end{split} 
\label{eq:vanilla}
\end{equation}
However, to address stability issues during training of the VGAN, the least squares GAN (LSGAN) was introduced \cite{lsgan}. In a conditional setting, it optimises the objective functions
\begin{equation}
\begin{split}
\min_{\D}\mathcal{L}_{LSGAN}(\D)  =  & \frac{1}{2}\mathbb{E}_{\boldsymbol{X},\boldsymbol{Y}}[(\D(\boldsymbol{X},\boldsymbol{Y})-b)^2]  + \\
& \frac{1}{2}\mathbb{E}_{\boldsymbol{X}}[( \D(\boldsymbol{X},\G(\boldsymbol{X}))-a)^2]\,,
\end{split} 
\label{eq:lsgan}
\end{equation}
\begin{equation}
 \min_{\G}\mathcal{L}_{LSGAN}(\G)  =  \frac{1}{2}\mathbb{E}_{\boldsymbol{X}}[( \D(\boldsymbol{X},\G(\boldsymbol{X}))-c)^2],
\label{eq:lsgan-gen}
\end{equation}
where $\boldsymbol{X}$ and $\boldsymbol{Y}$ are image patches from the input and the target domain, $a$ and $b$ are labels for false and real data, while $c$ denotes a value that $\G$ tricks $\D$ to believe for false data \cite{lsgan}. 

Isola \emph{et al.} \cite{isola} suggest to combine a GAN loss with an \L\ loss to reduce visual artefacts in the generated images. The contribution of the \L\ loss to the overall objective function is weighted by a regularisation parameter $\alpha$, which is determined by hyperparameter tuning. In \cite{s.bjorkPotentialSequentialNonsequential2022}, the LSGAN model was found to outperform a VGAN and a Wasserstein GAN \cite{wgangp}. We therefore replace \eqref{eq:lsgan-gen} with the following objective function for the generator of the baseline cGAN U-Net:
\begin{equation}
  \mathcal{L}_{cGAN}(\G)  = \mathcal{L}_{L1} + \alpha \mathcal{L}_{LSGAN}(\G), \quad \alpha\in[0,1].
    \label{eq:lsganl1}
\end{equation}
We find an optimal value of $\alpha=0.01$, as in \cite{isola}. 
Similar to \cite{isola}, we do not change the objective function of \D\ 
for the baseline cGAN U-Net. 

In \cite{isola}, different architectures were evaluated for the discriminator \D\ by altering the patch size $N$ of the receptive fields, ranging from a $1 \times 1$ PixelGAN to an $N \times N$ PatchGAN. The \D\ network applies convolutional processing to the pair of image patches to produce several classification responses, which are then averaged to determine whether the pair of image patches is real or false. In a PixelGAN, the discriminator attempts to classify each $1 \times 1$ pixel within the image patch as either real or false. In contrast, for the two PatchGAN networks, the discriminator tries to differentiate each $N \times N$ patch of pixels in the image patch as real or false. 

\subsection{Fine-tuning Stage}
\label{sec:finetune}
This section describes the loss functions used to fine-tune the PAR U-Net model and the cGAN U-Net model.
\subsubsection{Pixel- and frequency-aware regression U-Net}
To enforce the regression U-Net to focus on the alignment of the image frequency components during training, we propose to add a frequency-aware loss to the pixel-aware regression model or the adversarial cGAN model. We choose to employ the FFT loss from \cite{s.bjorkSimplerBetterSpectral2022}, which has shown promising results and is complementary to existing spatial losses. It is formulated as
\begin{equation}
\begin{split}
    \mathcal{L}_{FFT} = & \frac{1}{k}\sum \left(imag[\mathcal{F}(\boldsymbol{Y})]- imag[\mathcal{F}(\boldsymbol{\hat{Y}})]\right)^2 + 
    \\
    &  \frac{1}{k}\sum \left(real[\mathcal{F}(\boldsymbol{Y})]- real[\mathcal{F}(\boldsymbol{\hat{Y}})]\right)^2, 
    \end{split}
 \label{eq:fft}
\end{equation}
where $\mathcal F$ denotes the fast Fourier transform. The \lfft\ uses MSE to enforce alignment of the \textit{real} and \textit{imaginary} parts of target and generated image patches in the frequency domain.

The total composite loss function becomes:
\begin{equation}
\begin{split}
  \mathcal{L}_{tot} 
  &= \mathcal{L}_1 + \alpha \mathcal{L}_{LSGAN} + \gamma \mathcal{L}_{FFT}, \quad \alpha, \gamma\in[0,1]. 
     \label{eq:ltot}
\end{split}
\end{equation}
A regularisation parameter $\gamma$ is associated with \lfft\ to adjust its influence on \ltot. The $\alpha$ is still associated with \lcgan. All objective functions we use in the fine-tuning can be formulated with $\mathcal{L}_{tot}$, as we can ablate it by setting $\alpha=0$ or $\gamma=0$.

The baseline PAR U-Net model is either fine-tuned on the \lcgan\ loss (\ltot\ with $\gamma=0$), the combined \L\ and \lfft\ loss (\ltot\ with $\alpha=0$), or the \ltot\ loss. The baseline cGAN U-Net model is fine-tuned on \ltot. We refer to Appendix \ref{app:hyp} for an extensive evaluation of model settings and hyperparameters used in the pretraining and fine-tuning phase.

\subsection{Masked Loss Computation on Discontinuous Data}
Due to the discontinuity of the ALS-derived SV prediction maps from Norway, there is not a target pixel for every pixel of the continuous SAR predictor dataset. To remedy this, we introduce masked loss computation. In this way, the convolutional processing of the predictor data creates a wall-to-wall map of model predictions, but in the comparison with the target dataset, pixels without prediction targets are masked out and excluded from the learning process. In addition to masking the pseudo-targets, we want to boost learning for pixels and patches with true prediction targets, hence reducing the impact of pseudo-targets relative to true targets. 

As shown in \figr{fig:pipeline}, the training dataset contains two binary masks of the same size as the input and target data patches: the ground reference mask, \Mimp, and the pseudo-target mask, \Mf, which for the Tanzanian AOI contains only ones. Masked losses are computed through simple Hadamard products, i.e.\ element-wise multiplication, denoted $\odot$. For instance, the masked \L\ loss becomes:
\begin{equation}
\begin{split}
   \mathcal{L}_{1}^{\mathcal{M}} &= \mathcal{L}(\mathbfcal{M}\odot\boldsymbol{Y}, \mathbfcal{M}\odot\boldsymbol{\hat{Y}}) \\
   &= \frac{1}{N\times N}\sum_{i,j}{(\mathbfcal{M}\times|y_{i,j} - \hat{y}}_{i,j}|),
    \label{eq:maskl1}
\end{split}
\end{equation}
where $\mathbfcal{M}$ can be \Mf\ or \Mimp, $y_{i,j}$ and $\hat{y}_{i,j}$ are pixels of the target patch $\mathbf{Y}$ and the predicted target patch $\hat{\mathbf{Y}}$, whose size is $N\times N$. Similarly, $\mathcal{L}_{FFT}$ can be computed on $\mathcal{F(\mathbfcal{M}\odot\mathbfcal{Y})}$ and $\mathcal{F(\mathbfcal{M}\odot\hat{\mathbfcal{Y}})}$. Also the discriminator \D\ can be fed with masked patches, either the real pair $(\mathbfcal{M}\!\odot\!\boldsymbol{X},\mathbfcal{M}\!\odot\!\boldsymbol{Y})$ or the fake pair $(\mathbfcal{M}\odot\boldsymbol{X},\mathbfcal{M}\odot\G(\boldsymbol{X})$. With this input to \D, the \lsgan\ losses in \eqref{eq:lsgan} and \eqref{eq:lsgan-gen} generalise to the masked case.

Let loss functions masked with \Mf\ and \Mimp\ be denoted $\mathcal{L}^{\mathcal{M}_{pt}}$ and $\mathcal{L}^{\mathcal{M}_{gr}}$, respectively. To weight the true targets and the pseudo-targets differently, the total loss is decomposed as:
\begin{equation}
\begin{split}
\label{eq:masked-loss}
    \mathcal{L}_{tot} &= \delta \mathcal{L}_{tot}^{\mathcal{M}_{gr}} + \mathcal{L}_{tot}^{\mathcal{M}_{pt}} \\
    &= \delta \mathcal{L}_{1}^{\mathcal{M}_{gr}} + \mathcal{L}_{1}^{\mathcal{M}_{pt}}  + \gamma\left(\delta\mathcal{L}_{FFT}^{\mathcal{M}_{gr}} + \mathcal{L}_{FFT}^{\mathcal{M}_{pt}}\right)\\
    &\;+ \alpha\left(\delta\mathcal{L}_{LSGAN}^{\mathcal{M}_{gr}} + \mathcal{L}_{LSGAN}^{\mathcal{M}_{pt}}\right)\,,
    \end{split}
\end{equation}
with  $\alpha, \gamma\in[0,1]$ and true target weighting parameter $\delta\gg 1$,  found from hyperparameter tuning (see Appendix \ref{app:hyp}). A masked loss decreases when the mask has many zeros, which is as intended, since the amount of true or pseudo-targets contained in a patch should determine its impact.
This is inspired by pseudo-labelling \cite{leePseudolabelSimpleEfficient2013}, a related semi-supervised learning algorithm for categorical prediction. It recommends to balance the losses computed over pseudo-labels (the categorical equivalent to the pseudo-targets in the regression task) and true labels, as there are generally much more pseudo-labels than true labels. In our training paradigm, this translates to boosting the masked loss computed over the true targets.

\begin{table*}[]
\caption{Quantitative evaluation of models trained on the Tanzanian dataset. Metrics of $RMSE$ and $MAE$ are measured with respect to the pseudo-target data (left side of the table) and ground reference data (right side of the table). CV-RMSE is given as mean $\pm$ standard deviation. Numbers in boldface indicate the best-performing model per column. All units are in $\mathrm{Mg\,ha}^{-1}$}    
\label{tab:restanz}
    \centering
    \begin{tabular}{l l |c c c | l c  l }
         \hline
   \multicolumn{2}{c|}{\textbf{Models}} & \textbf{RMSE} $\downarrow$&\textbf{CV-RMSE} $\downarrow$& \textbf{MAE} $\downarrow$ & \textbf{RMSE} $\downarrow$  &\textbf{CV-RMSE} $\downarrow$& \textbf{MAE} $\downarrow$\\
 \textbf{Pretraining:} & \textbf{Fine-tuning:} &  \multicolumn{3}{c|}{w.r.t.\ pseudo-target dataset} & \multicolumn{3}{c}{w.r.t.\ ground reference data} \\

   \hline 
   Baseline ALS$^a$    & \quad \quad -- & -- & -- &  --  & 33.39 & ---  & 24.61\\
   Sequential cGAN$^{b}$ & \quad \quad  -- & -- &  -- & --  & 39.84 &   --- & 31.46\\
   \hdashline
  PAR U-Net (\L)  &\quad \quad   -- & 29.82  & 29.47 $\pm$ 0.15   &  20.05  & 34.24 & 35.64 $\pm$ 7.31& 25.82\\
  \multicolumn{2}{l|}{cGAN U-Net (\lcgan) \qquad\;\;\; --} &  32.02  & 31.19 $\pm$ 0.21  & 22.46  & 37.22 & 38.78 $\pm$ 5.9 & 28.91\\
  \hline
  PAR U-Net (\L)& \lcgan\ & 29.31 & 29.81 $\pm$ 0.51   & 19.92  & \textbf{32.91}$^{\bullet}$ & \textbf{35.53 $\pm$ 4.14} &  \textbf{25.53} \\
  PAR U-Net (\L) &  \L\ + \lfft & \textbf{26.10}   &\textbf{ 26.17 $\pm$ 0.05} &   \textbf{18.11} & 34.40 & 36.13 $\pm$ 3.09  & 26.54\\
  PAR U-Net (\L) & \lcgan\ + \lfft  & 32.75 & 33.58 $\pm$ 0.54 & 24.21  & 36.19 & 37.46 $\pm$ 5.36&28.24\\
  cGAN U-Net (\lcgan)\ & \lcgan\ + \lfft  & 29.40 & 30.73 $\pm$  0.26   &  20.85 &37.65 & 38.49 $\pm$ 7.17 & 29.18 \\
    \hline 
    \multicolumn{7}{l}{}\\
    \multicolumn{7}{l}{${}^\text{a}$ The conventional ALS-based statistical regression model proposed in \cite{naessetMappingEstimatingForest2016}. Metrics are retrieved from \cite{naessetMappingEstimatingForest2016} and \cite{s.bjorkPotentialSequentialNonsequential2022}.} \\
    \multicolumn{7}{l}{${}^\text{b}$ The optimal cGAN-based sequential regression models proposed in \cite{s.bjorkPotentialSequentialNonsequential2022}. Provided metrics are from the same source.} \\
    \multicolumn{7}{l}{${}^\text{$\bullet$}$ Indicates that a model performs better than the Baseline ALS model.} \\
    \end{tabular}
\end{table*} 
 
\section{Experimental Results}\label{sec:res}
This section presents experimental results of the prediction models trained on the Tanzanian and Norwegian datasets. We provide results on both regional and pan-regional models for the Norwegian datasets. The pan-regional models have been trained on all available training datasets from \nl, \tyr\ and \hole. The regional models were trained on datasets from either \nl, \tyr\ or \hole, and evaluated on the test data from the same region it was trained on. Appendix \ref{app:hyp} provides details on hyperparameter tuning and settings used during model training. 

Results are given both for the pretraining stage, i.e.\ the baseline models, and the fine-tuning stage as root mean square error (RMSE) and mean absolute error (MAE). Models with a low RMSE and MAE are preferred, as indicated by the symbol $\downarrow$ in the tables. Models have been trained to in two ways: (i) using all true target imputed with pseudo-targets; (ii) in cross-validation (CV) mode 
by rotationally imputing 80\% of the target labels with the available pseudo-targets. For the latter case, a CV-RMSE is reported as $\mu$ (mean) $\pm \sigma$ (standard deviation).
In the evaluation, we report model performance on the true targets and unseen test dataset. Since the CNN models work on image patches, model predictions are inferred by processing the AOI as $64 \times 64$ \sen image patches with 50\% overlap. A wall-to-wall prediction map is created by mosaicking patches through linear image blending, using the p-norm with a heuristic value of $p\!=\!5$, as proposed in \cite{s.bjorkPotentialSequentialNonsequential2022}. 

\subsection{Results: Tanzania models} \label{sec:restz}
The Tanzanian test set consists of 14 patches of pseudo-target AGB predictions and true targets from the 88 field plots. Quantitative results in terms of model performance on both the pseudo-target dataset and on the true targets are given in \tabref{tab:restanz}. Metrics for the original ALS-derived AGB model, see \cite{naessetMappingEstimatingForest2016, s.bjorkPotentialSequentialNonsequential2022}, and the best sequential cGAN model from \cite{s.bjorkPotentialSequentialNonsequential2022} are also provided. Note that the best cGAN model from \cite{s.bjorkPotentialSequentialNonsequential2022} was trained only on pseudo-targets, without access to true targets. We do not report the performance of the original ALS-derived AGB model and the sequential cGAN model on the test dataset, or the $\mu \pm \sigma$ CV-RMSE on the true targets, as these metrics were not provided in \cite{naessetMappingEstimatingForest2016, s.bjorkPotentialSequentialNonsequential2022}. All units in \tabref{tab:restanz} are of $\mathrm{Mg\,ha}^{-1}$. Numbers in boldface indicate the best-performing model per column, while ($\bullet$) indicates that a model performs better than the baseline ALS model.
 \subsection{Results: Norwegian models} \label{sec:resnor}

\begin{table*}[]
\caption{Quantitative evaluation of pan-regional Norwegian models. Metrics for each region are provided with respect to pseudo-target data (left side of the table) and ground reference data (right side of the table) as RMSE, MAE and CV-RMSE, the latter given as mean $\pm$ standard deviation. Numbers in boldface indicate the best-performing model per column. All units are in $m^3ha^{-1}$}
    \label{tab:resnatnor}
    \centering
    \begin{tabular}{l l | c c c | l c  l}
         \hline
   \multicolumn{2}{c|}{\textbf{Models}} & \textbf{RMSE} $\downarrow$&\textbf{CV-RMSE} $\downarrow$& \multicolumn{1}{c|}{\textbf{MAE} $\downarrow$}& \textbf{RMSE} $\downarrow$  &\textbf{CV-RMSE} $\downarrow$ & \textbf{MAE} $\downarrow$\\
 \textbf{Pretraining:} & \textbf{Fine-tuning:} & \multicolumn{3}{c|}{w.r.t.\ pseudo-target data} & \multicolumn{3}{c}{w.r.t.\ ground reference data} \\
 \hline
   \multicolumn{8}{c}{\textbf{Region: Nordre Land}}\\ 
   \hline
 Baseline ALS & \quad \quad -- & --  & -- & --  & 83.54 & -- & 63.29 \\
 \hdashline
 PAR U-Net (\L)  &   \quad \quad -- & \textbf{68.08} & \textbf{68.67 $\pm$ 0.29}  &  \textbf{31.89} &  73.72$^{\bullet}$ &  92.63 $\pm$ 3.74 & 38.32$^{\bullet}$ \\
  \multicolumn{2}{l|}{cGAN U-Net (\lcgan) \qquad\;\;\; --} & 77.66 & 77.22 $\pm$ 3.30 &  34.31 & 88.30 & 120.6 $\pm$ 27.35 & 48.03$^{\bullet}$ \\
  \hline
  PAR U-Net (\L) & \lcgan\ &  68.94 &69.17  $\pm$ 0.26 & 32.90 & 73.93$^{\bullet}$  &  \textbf{79.36$\pm$  3.37}  &  \textbf{32.86}$^{\bullet}$ \\
  PAR U-Net (\L) & \L\ + \lfft    & 72.38 &  71.96 $\pm$ 0.48  & 35.32 & 70.92$^{\bullet}$ & 79.66 $\pm$ 4.01  & 33.66$^{\bullet}$\\
PAR U-Net (\L) & \lcgan\ + \lfft &  71.57 & 72.84 $\pm$ 2.34 & 34.03 &  \textbf{68.72}$^{\bullet}$  & 99.67 $\pm$ 36.23 &  37.71$^{\bullet}$ \\
  cGAN U-Net (\lcgan)\ & \lcgan\ + \lfft  & 78.48 & 77.22 $\pm$ 1.19  & 37.46  &  93.09  & 92.55  $\pm$ 33.16 & 53.51$^{\bullet}$\\
    \hline 
   \multicolumn{8}{c}{\textbf{Region: Tyristrand}}\\ 
   \hline
   Baseline ALS  & \quad \quad --  & -- & -- & --  & 75.62  & -- & 59.17\\
   \hdashline
  PAR U-Net (\L) & \quad \quad --  & \textbf{62.31} & \textbf{63.01 $\pm$ 1.37} &  \textbf{24.83}  & 43.77$^{\bullet}$  & 58.96 $\pm$ 6.20 & \textbf{28.30}$^{\bullet}$\\
  \multicolumn{2}{l|}{cGAN U-Net (\lcgan) \qquad\;\;\; --} & 84.96 & 79.43 $\pm$ 6.44 & 30.17 &  76.59 & 98.98 $\pm$ 14.35 & 55.68$^{\bullet}$ \\
  \hline
  PAR U-Net (\L) & \lcgan\ & 64.22 & 66.51 $\pm$  0.93  &   25.77 &  \textbf{40.75}$^{\bullet}$  & \textbf{45.78 $\pm$ 4.61}  & 28.56$^{\bullet}$\\
PAR U-Net (\L) &\L\ + \lfft      & 76.86 &  75.63 $\pm$ 1.41 & 28.94 &  42.80$^{\bullet}$  &  49.17 $\pm$ 2.16 &  28.28$^{\bullet}$\\
 PAR U-Net (\L) & \lcgan\ + \lfft & 65.50 &  65.99 $\pm$  1.72 & 26.74  & 55.04$^{\bullet}$ & 74.89 $\pm$ 22.09  & 35.74$^{\bullet}$ \\
  cGAN U-Net (\lcgan)\ & \lcgan\ + \lfft  & 84.84 &  78.30 $\pm$ 5.95  &  29.88 & 101.55  & 98.54 $\pm$ 14.62 & 65.96\\
    \hline 

   \multicolumn{8}{c}{\textbf{Region: Hole}}\\ 
   \hline
Baseline ALS &  \quad \quad -- & -- & -- & --  &   60.94 & -- & 50.06 \\
\hdashline
   PAR U-Net (\L) &  \quad \quad --  & \textbf{113.82}  & \textbf{116.18 $\pm$ 1.00}  & \textbf{57.68}  & 72.47 & 82.43  $\pm$ 7.37 & 41.39$^{\bullet}$\\
 \multicolumn{2}{l|}{cGAN U-Net (\lcgan) \qquad\;\;\; --} & 136.03 &  129.57 $\pm$ 3.62 & 65.24  & 95.61 & 126.87 $\pm$ 12.53  & 64.37 \\
 \hline
   PAR U-Net (\L) & \lcgan\ & 121.14 &  124.71 $\pm$ 1.15   & 59.90 & 67.54 & 72.24$\pm$ 5.49 & \textbf{37.05}$^{\bullet}$\\
PAR U-Net (\L) & \L\ + \lfft & 124.13 &  123.51 $\pm$ 0.54 &  61.09  &  \textbf{64.69}  & \textbf{69.89 $\pm$ 3.13} & 39.22$^{\bullet}$\\
 PAR U-Net (\L) & \lcgan\ + \lfft & 114.59 & 116.10 $\pm$ 1.78 & 59.13  & 68.83  & 94.44 $\pm$ 27.35 & 43.06$^{\bullet}$\\
  cGAN U-Net (\lcgan)\ & \lcgan\ + \lfft & 123.98 & 125.40 $\pm$ 2.39 &  60.99 & 94.10  & 122.90 $\pm$ 17.08  &59.66\\
    \hline 
\multicolumn{7}{l}{${}^\text{$\bullet$}$ Indicates that a model performs better than the Baseline ALS model.} \\
    \end{tabular}
\end{table*}

 The Norwegian models have all been trained to translate \sen data into ALS-derived SV predictions for commercial forests. \tabref{tab:patches} shows that data from \nl\ is over-represented in the Norwegian dataset. I.e., approximately 80\% of the training image patches are from \nl, while only 6\% are from the \hole\ region. Four types of Norwegian models were developed: one pan-regional model that represents all three regions and separate regional models for \nl, \tyr\ and \hole. The pan-regional models were trained on pooled training data from all regions, but evaluated separately on each region's pseudo-target data and true target data. The three regional models were both trained and evaluated on data from each separate region. 
 
Since \nl\ is over-represented in the dataset, we wish to investigate if the pan-regional models evaluated on \nl\ perform similarly to the corresponding regional models developed for \nl.
On the other hand, as the available data from both \hole\ and \tyr\ are limited, we wish to compare the respective regional models to the pan-regional model. The aim is to identify and quantify any difference in performance and, if possible, to draw conclusions about transferability and impacts of dataset size. 

As for the Tanzania, different CNN models were evaluated against each other by comparing their performance on unseen test patches of pseudo-target data and on true targets of field measured SV. The number of field plots, i.e.\ true targets, in each region, can be found in \tabref{tab:regions}. The Hole test set consists of 14 patches of pseudo-target data, Tyristrand of 25 and Nordre Land of 87 test patches, each of $64 \times 64$ pixels. 

Quantitative results from the evaluation of the pan-regional Norwegian models are listed in \tabref{tab:resnatnor} while \tabref{tab:reslocnor} lists results for the regional models. For the regional models, only results for the baseline PAR U-Net model and the model pretrained on \L\ and fine-tuned with the \lcgan\ loss are given, as these have proven to be robust on both the Tanzanian data and the pan-regional Norwegian dataset. Metrics obtained with the original ALS-derived SV model have been computed for each region by extracting the area-weighted mean of ALS-derived SV predictions at the location of each field plot. The CV-RMSE for the original ALS-derived SV models were not provided to us for this work and are therefore not given in  \tabref{tab:resnatnor} or \tabref{tab:reslocnor}. All metrics in both tables are in units of $\mathrm{m}^3\,\mathrm{ha}^{-1}$. Boldface numbers in a column of \tabref{tab:restanz} indicate the model that performs best. A ($\bullet$) symbol indicates that a model performs better than the baseline ALS model.

\begin{table*}[]
\caption{Quantitative evaluation of regional Norwegian models. Metrics for each region are provided with respect to pseudo-target data (left side of the table) and ground reference data (right side of the table) as RMSE, MAE and CV-RMSE, the latter given as mean $\pm$ standard deviation. Numbers in boldface indicate the best-performing model per column. All units are of $m^3ha^{-1}$.}
    \label{tab:reslocnor}
    \centering
    \begin{tabular}{l l | c c  c | l c  l}
         \hline
   \multicolumn{2}{c|}{\textbf{Models}} & \textbf{RMSE} $\downarrow$&\textbf{CV-RMSE} $\downarrow$& \multicolumn{1}{c|}{\textbf{MAE} $\downarrow$}& \textbf{RMSE} $\downarrow$  &\textbf{CV-RMSE} $\downarrow$& \textbf{MAE} $\downarrow$\\
 \textbf{Pretraining:} & \textbf{Fine-tuning:} & \multicolumn{3}{c|}{w.r.t.\ pseudo-target data} & \multicolumn{3}{c}{w.r.t.\ ground reference data}\\
 \hline
   \multicolumn{8}{c}{\textbf{Region - Nordre Land: }}\\ 
   \hline
 Baseline ALS & \quad \quad --  & -- & -- & --  & 83.54 & --  & 63.29 \\
 \hdashline
PAR U-Net (\L) & \quad \quad --  & \textbf{69.65} & \textbf{69.48$\pm$ 0.23}   &  \textbf{31.89}  & 75.53$^{\bullet}$  &  92.36 $\pm$ 2.06 & 36.82$^{\bullet}$\\
  PAR U-Net (\L) & \lcgan\ & 70.45 & 70.51 $\pm$  0.15  & 32.94  &  \textbf{69.41}$^{\bullet}$  & \textbf{82.40 $\pm$ 3.68} & \textbf{32.70$^{\bullet}$}\\
    \hline 
   \multicolumn{8}{c}{\textbf{Region - Tyristrand: }}\\ 
   \hline
   Baseline ALS & \quad \quad -- & -- & --  & --  & 75.62  & -- & 59.17\\
   \hdashline
  PAR U-Net (\L) & \quad \quad --  & \textbf{66.77} & \textbf{67.96 $\pm$ 0.59}  &  \textbf{27.79}  & 48.04$^{\bullet}$ & 63.77$\pm$ 8.22 & 32.5$^{\bullet}$ \\
    PAR U-Net (\L) & \lcgan\  & 68.80 & 69.36 $\pm$ 0.42  & 28.96  &  \textbf{45.22}$^{\bullet}$ & \textbf{60.73 $\pm$ 4.80} & \textbf{27.35$^{\bullet}$} \\
    \hline 
   \multicolumn{8}{c}{\textbf{Region - Hole:}}\\ 
   \hline
Baseline ALS & \quad \quad --  & -- & -- & -- &  60.94 & -- &  50.06 \\
\hdashline
PAR U-Net (\L) & \quad \quad --  & 136.19 &  138.82 $\pm$   1.13  &  71.88 &74.60 &  99.48 $\pm$ 10.47 & 42.55$^{\bullet}$ \\
 PAR U-Net (\L) & \lcgan & \textbf{132.47} & \textbf{132.58 $\pm$  0.76}  &  \textbf{70.73}  &  \textbf{65.47} & \textbf{77.38$\pm$ 5.37}  & \textbf{38.22}$^{\bullet}$\\
    \hline 
    \multicolumn{8}{l}{${}^\text{$\bullet$}$ Indicates that a model performs better than the Baseline ALS model.} \\
    \end{tabular}
\end{table*}

\section{Discussion} \label{sec:disc}
Six new CNN-based regression models (two baseline and four fine-tuned ones) have been developed to improve earlier work on the Tanzanian dataset using the semi-supervised imputation strategy proposed herein. Above all, \tabref{tab:restanz} shows that the model pretrained on the \L\ loss and fine-tuned on the \lcgan\ loss performs better than the conventional statistical ALS-based AGB model proposed in \cite{naessetMappingEstimatingForest2016}, and all other Tanzanian models on the field data. The CNN model that most accurately recreates the AGB pseudo-target data is pretrained on the \L\ loss and fine-tuned on the combined \L\ and \lfft\ loss, see \tabref{tab:restanz}. The results on the Tanzanian dataset show the potential of a two-stage training paradigm and of frequency-aware training to reduce the impact of spectral bias. 
Furthermore, the results in \tabref{tab:restanz} show that the baseline PAR U-Net model performs better than the baseline cGAN U-Net model on both the pseudo-target and the true target data. These findings align with existing knowledge in the field of image super-resolution: It is disadvantageous to adopt a purely adversarial training strategy on tasks that require high reconstruction accuracy in terms of RMSE. In this case, employing a simpler pixel-wise regression U-Net is better. The proposed baseline cGAN U-Net model is most similar to the sequential cGAN model proposed in \cite{s.bjorkPotentialSequentialNonsequential2022}. \tabref{tab:restanz} shows that the proposed semi-supervised imputation strategy improves the CNN models' performance in AGB prediction. 

Several new CNN models are also proposed for SV prediction on the Norwegian datasets. Our approach is to train pan-regional models by combining data from all three Norwegian regions, \nl, \tyr\ and \hole, followed by evaluation of test and field data from each individual region. The purpose of the pan-regional models is to develop models that generalise well to more than one region, which is particularly advantageous for regions with little training data. As a result, these models hold the potential for substantial cost-savings if field work can be reduced during operational inventories. 

According to \tabref{tab:resnatnor}, the baseline PAR U-Net model outperforms the other models in accurately recreating the pseudo-target SV data. We advise to avoid the baseline cGAN U-Net model or the pan-regional model that was pretrained on the \lcgan\ loss, followed by fine-tuning on the combined \lcgan\ and \lfft\ losses, when training CNN models for SV prediction. As the models are evaluated on RMSE and not perceptual quality, the results suggest that adversarial training should be avoided in the initial training phase.
As demonstrated in \tabref{tab:resnatnor}, fine-tuning and the composition of losses generally improve model performance with respect to field data, with few exceptions. Moreover, all pan-regional fine-tuned models perform better than the conventional statistical ALS-based models derived for either \nl, \tyr, or \hole. Based on the CV-RMSE, we recommend using fine-tuned models that are pretrained on \L\ and fine-tuned on either the combined \L\ and \lfft\ loss or on \lcgan. For instance, the model fine-tuned on the combined \L\ and \lfft\ loss performs best on the Hole field data, whereas the model fine-tuned on \lcgan\ performs best on Tyristrand field data. 

In addition to the pan-regional models trained on the whole Norwegian dataset, regional models were developed for this work. Unlike the pan-regional models, these were only trained and evaluated on a specific region. \tabref{tab:patches} shows a significant difference in the amount of available training data among the regions. The \hole\ region has the least data, followed by \tyr, while \nl\ has the most data. Consequently, it implies that the regional \nl\ model has been trained on almost the same training data as the pan-regional model. For \hole\ (and \tyr), the regional models are trained on only a fraction of the training data available for the pan-regional model, which could impact their relative performance. 
Based on the discussion above, we train the following two models: a regional PAR U-Net model and a regional model pretrained on \L\ and fine-tuned on the \lcgan\ loss. The fine-tuned model was chosen among the other three, as it has proven to be robust on both the Tanzanian data and the pan-regional Norwegian dataset.

In general, comparing the results of the pan-regional models in \tabref{tab:resnatnor} to the regional models in \tabref{tab:reslocnor}, we observe that the pan-regional Norwegian models perform better than all regional models with one exception. The regional \nl\ model pretrained on the \L\ objective and fine-tuned on the \lcgan\ objective performs better than the pan-regional model on the corresponding regional model on field data. These results show the potential of training regional models that utilises all available data from nearby regions. 

To our knowledge, it is the first time that the \lfft\ loss has been evaluated outside the natural image domain, e.g.\ on remote sensing images. Our results from both the Tanzanian and Norwegian models show that the simple \lfft\ objective function efficiently reduces the impact of spectral bias and thereby improves the performance of the CNN model. 

\section{Conclusion}\label{sec:conc}
Through the use of a semi-supervised imputation strategy, we demonstrate the ability of contextual generative CNN models to accurately map \sen C-band data to target data consisting of spatially disjoint polygons of ALS-derived prediction maps. The generalisation ability of our modelling approach was evaluated for AGB prediction in the Tanzanian miombo woodlands and for SV prediction in three managed boreal forests in Norway. Our results show that the models developed using the imputation strategy achieve state-of-the-art performance compared to previous studies, suggesting that the contextual C-band SAR-based models outperform conventional statistical ALS-based models in accurately predicting the target labels of ground reference data.
Furthermore, we demonstrate that a two-phased learning strategy, which includes pretraining with a pure pixel-wise regression U-Net followed by either a regression cGAN model or a pixel- and frequency-aware regression U-Net in the fine-tuning phase, improves model performance. We argue that pixel-aware pretraining enforces the model to focus on pixel-to-pixel relationships before learning general relationships.

\appendices
\vskip -2\baselineskip plus -1fil

\section*{Acknowledgements}
\label{sec:ack}
We gratefully acknowledge the Norwegian University of Life Sciences, the Tanzania Forest Services Agency, Prof.\ Eliakimu Zahabu and coworkers at Sokoine University of Agriculture, Viken Skog and the Swedish University of Agricultural Sciences for participation in field work and provision of in situ measurements, ALS-derived AGB and SV products. Special thanks to Prof.\ Håkan Olsson for providing ALS data acquired by SLU and to Mr.\ Svein Dypsund at Viken Skog for providing in situ measurements in Norway. Many thanks to Assoc.\ Prof.\ Benjamin Ricaud for valuable input on relevant experience from the field of single-image super-resolution.

\ifCLASSOPTIONcaptionsoff
  \newpage
\fi


\begin{IEEEbiography}[{\includegraphics[width=1in,height=1.25in,clip,keepaspectratio]{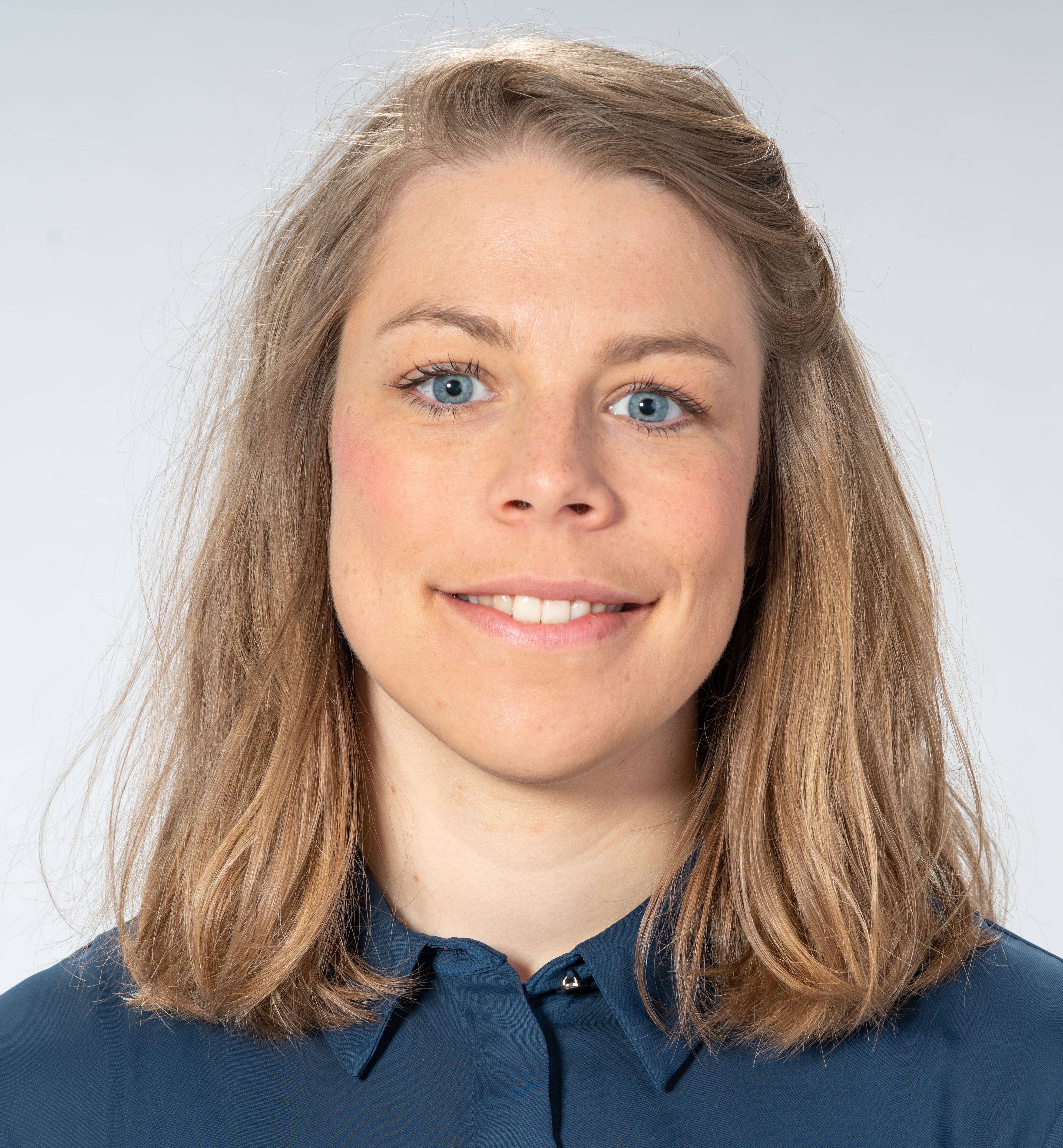}}]{Sara Björk}
received the M.Sc.\ degree in Applied Physics and Mathematics from UiT The Arctic University of Norway, in 2016, where she is currently pursuing the Ph.D. degree in physics. Since 2022, she has been working as a system developer in the Earth Observation Team at KSAT Kongsberg Satellite Services. Her research interests include computer vision, image processing, and deep learning, with a particular focus on developing methodologies that leverage deep learning techniques and remote sensing data for forest parameter retrieval.
\end{IEEEbiography}

\vskip -2\baselineskip plus -1fil
\begin{IEEEbiography}[{\includegraphics[width=1in,height=1.25in,clip,keepaspectratio]{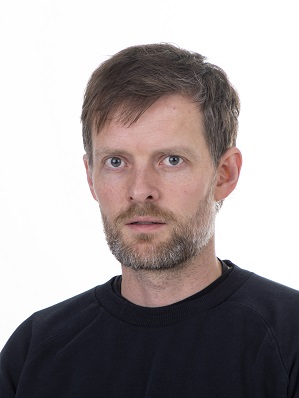}}]{Stian Normann Anfinsen}
received the M.Sc.\ degree in communications, control and digital signal processing from the University of Strathclyde, Glasgow, UK (1998) and the Cand.scient.\ (2000) and Ph.D.\ degrees (2010) in physics from UiT The Arctic University of Norway (UiT), Tromsø, Norway. He is a faculty member at the Dept.\ of Physics and Technology at UiT since 2014, currently as adjunct professor in machine learning. Since 2021 he is a senior researcher with NORCE Norwegian Research Centre in Troms\o. His research interests are in statistical modelling and machine learning for image and time series analysis.
\end{IEEEbiography}

\vskip -2\baselineskip plus -1fil
\begin{IEEEbiography}[{\includegraphics[width=1in,height=1.25in,clip,keepaspectratio]{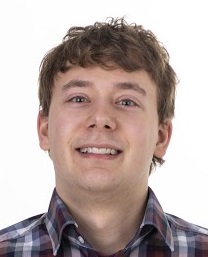}}]{Michael Kampffmeyer}
is an associate professor and head of the Machine Learning Group at UiT The Arctic University of Norway. He is also an adjunct senior research scientist at the Norwegian Computing Center in Oslo. His research interests include explainable AI and learning from limited labels (e.g.\ clustering, few/zero-shot learning, domain adaptation and self-supervised learning). Kampffmeyer received the Ph.D. degree from UiT in 2018. He has had long-term research stays in the Machine Learning Department at Carnegie Mellon University and Berlin Center for Machine Learning at the Technical University of Berlin. He is general chair of the annual Northern Lights Deep Learning Conference. 
\end{IEEEbiography}

\vskip -2\baselineskip plus -1fil
\begin{IEEEbiography}[{\includegraphics[width=1in,height=1.25in,clip,keepaspectratio]{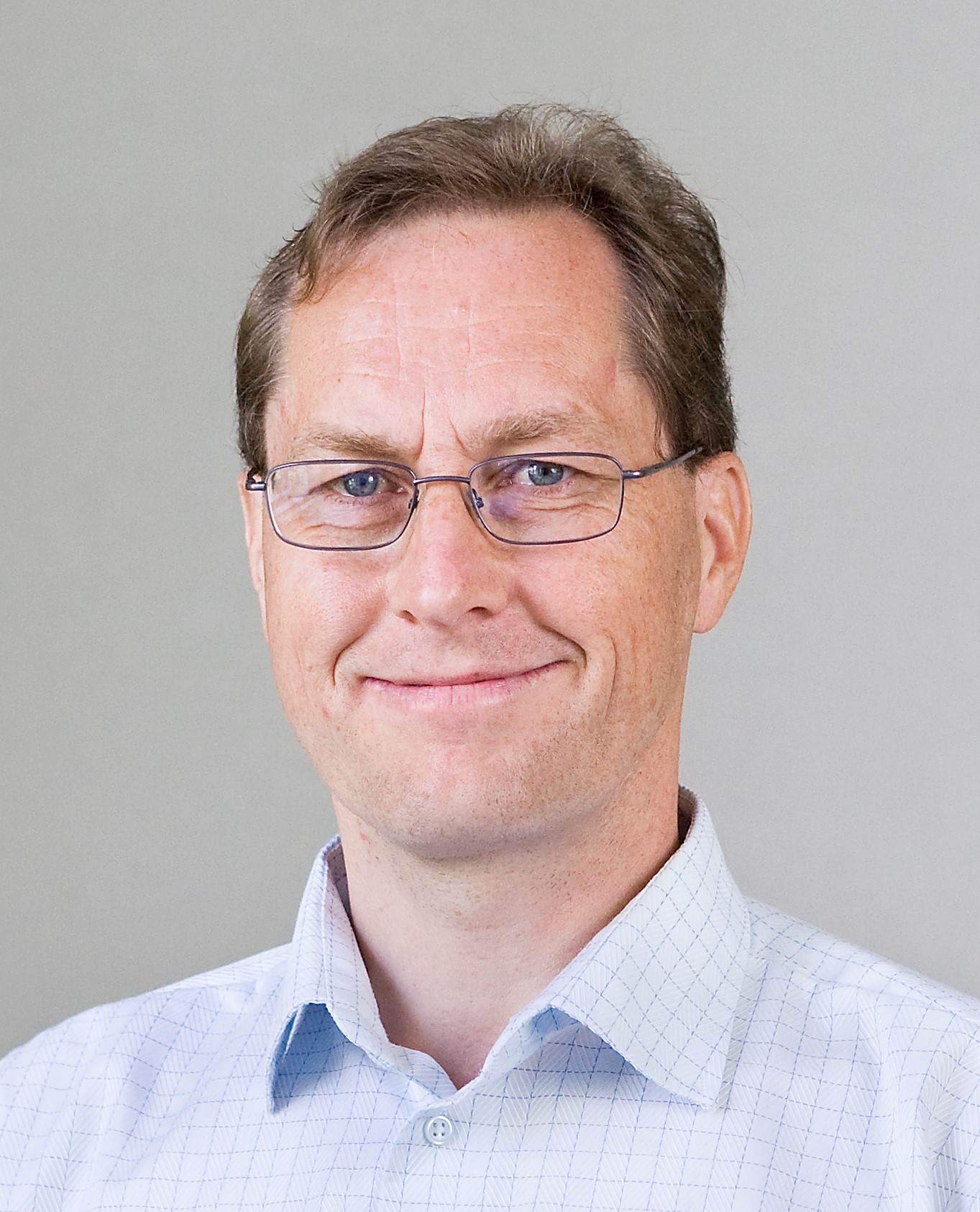}}]{Erik N\ae sset}
received the M.Sc. degree in forestry and the Ph.D.\ degree in forest inventory from the Agricultural University of Norway, Ås, Norway, in 1983 and 1992, respectively. His major field of research is forest inventory and remote sensing, with particular focus on operational management inventories, sample surveys, photogrammetry, and airborne LiDAR. He has played a major role in developing and implementing airborne LiDAR in operational forest inventory. He has been the leader and coordinator of more than 60 research programs funded by the Research Council of Norway, the European Union, and private forest industry. He has published around 250 papers in international peer-reviewed journals. His teaching includes lectures and courses in forest inventory, remote sensing, forest planning, and sampling techniques.
\end{IEEEbiography}

\vskip -2\baselineskip plus -1fil
\begin{IEEEbiography}[{\includegraphics[width=1in,height=1.25in,keepaspectratio]{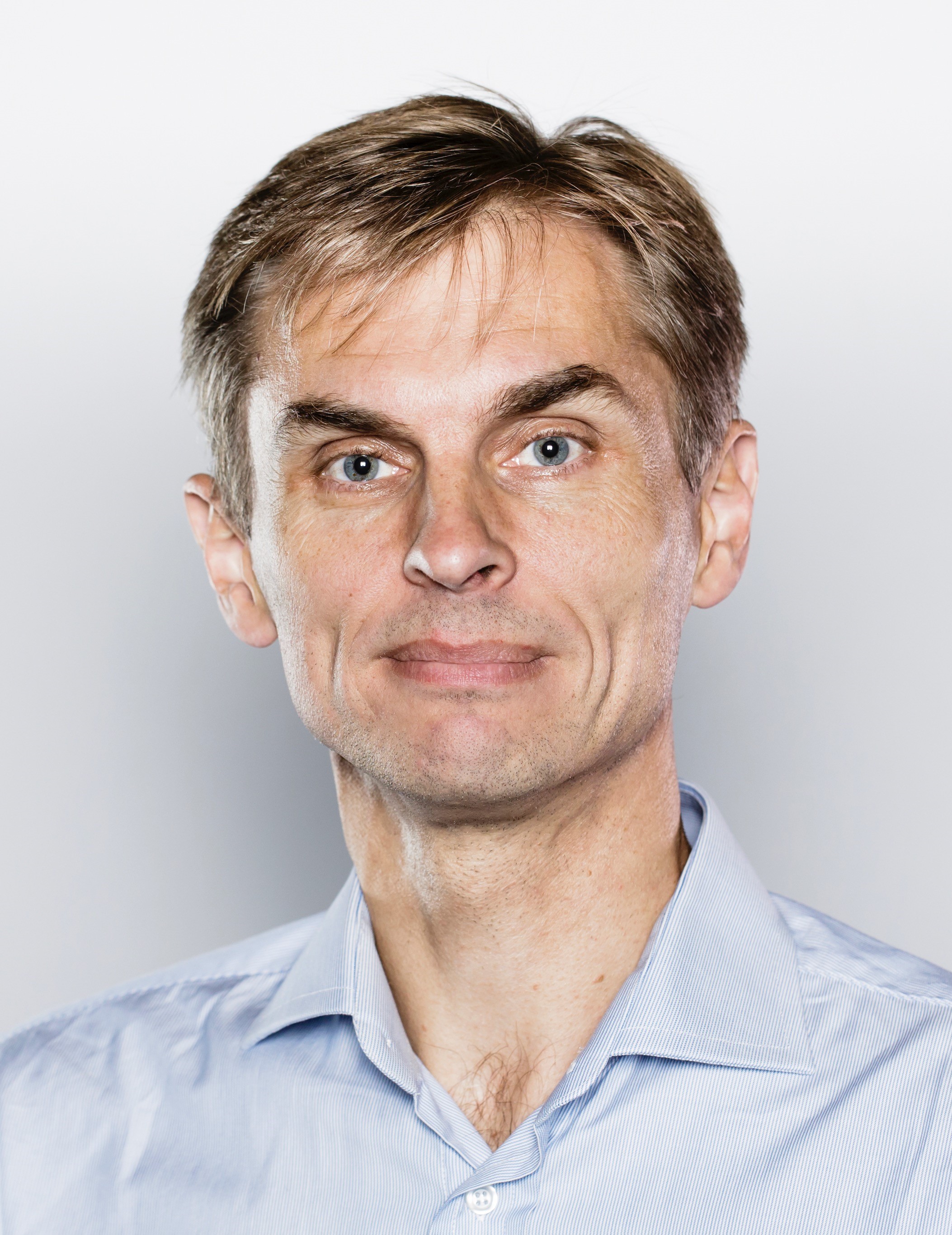}}]{Terje Gobakken}
is professor in forest planning and has published more than 190 peer-reviewed scientific articles related to forest inventory and planning in international journals. He has been working at the Norwegian National Forest Inventory and participated in compiling reports of emissions and removals of greenhouse gases from land use, land-use change and forestry in Norway. He has coordinated and participated in a number of externally funded projects, including international projects funded by for example NASA and EU, and has broad practical and research-based experience with development of big data and information infrastructures for forest inventory, planning and decision support.
\end{IEEEbiography}

\vskip -2\baselineskip plus -1fil
\begin{IEEEbiography}[{\includegraphics[width=1in,height=1.25in,clip,keepaspectratio]{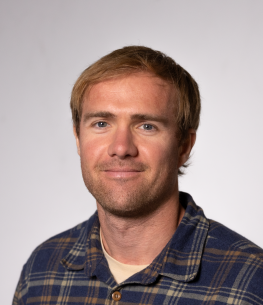}}]{Lennart Noordermeer}
received the M.Sc.\ degree in forestry and the Ph.D.\ degree in forest inventory from the Norwegian University of Life Sciences (NMBU) in 2017 and 2020, respectively. He currently has a researcher position in the Forest Inventory Group at the Faculty of Environmental Sciences and Natural Resource Management, NMBU. His research focuses on operational forest inventory, with emphasis on the use of data from forest harvesters as well as the use of multitemporal remotely sensed data for forest productivity estimation.
\end{IEEEbiography}

\bibliographystyle{IEEEtran}

\bibliography{references.bib}

\appendix
\section*{} \label{sec:app}
\begin{table*}[]
\caption{Search range for hyperparameters and model settings used in Tanzanian and Norwegian models.}
    \label{tab:hypbaselinerange}
    \centering
    \begin{tabular}{l l l }
         \hline
  Hyperparameters & Tanzanian dataset  & Norwegian dataset\\
   \hline
   \hline 
  Batch size (BS) & [2, 4, 6] & [8, 32, 64, 128] \\
  $\beta1$ (Adam optimiser) &  [0.4, 0.5, 0.6, 0.7, 0.8, 0.9] &  [0.4, 0.5, 0.6, 0.7, 0.8, 0.9] \\
  Learning rate (lr) & [1e-2, 1e-3, 2e-3, 1e-4, 2e-4, 2e-5, 1e-5]  & [1e-2, 1e-3, 2e-3, 1e-4, 2e-4, 2e-5, 1e-5] \\
  Encoder Network &  [ResNet18, ResNet34, ResNet50] & [ResNet18, ResNet34, ResNet50] \\
  Encoder, Decoder depth  & [4, 5]  & [4, 5] \\
  Discriminator network & [PixelGAN, PatchGAN(16), PatchGAN(34)] & [PixelGAN, PatchGAN(16), PatchGAN(34)]\\
  cGAN objective & [VGAN, LSGAN] & [VGAN, LSGAN] \\
Normalisation & [Instance, Batch, None] & [Instance, Batch, None] \\  
  $\alpha$ & [0.01, 0.1, 1] &  [0.01, 0.1, 1]\\
  $\gamma$ & [1e-8, 3e-8, 5e-8, 7e-8, 9e-8, 1e-7] & [1e-8, 3e-8, 5e-8, 7e-8, 9e-8, 1e-7] \\
  True target weight $\delta$ & [100, 200, 300, 400, 500] & [200, 300, 400, 500, 600, 700] \\
    \hline
  \end{tabular}
\end{table*}   

\subsection{Hyperparameter tuning for model selection}\label{app:hyp}

Extensive hyperparameter tuning was performed on the training and validation datasets for the Tanzanian and Norwegian models. Experiment tracking with Weights \& Biases sweeps \cite{wandb} was used, employing grid search during both pretraining and fine-tuning phases. Unlike most studies in forestry deep learning research \cite{hamedianfarDeepLearningForest2022}, the Adam optimiser was used for all proposed models. Hyperparameter tuning focused on finding optimal batch size (BS), $\beta_1$ value for the Adam optimiser, learning rate (lr), encoder network, encoder/decoder depth, discriminator network, cGAN loss, and number of epochs. Three discriminator networks were evaluated: $1 \times 1$ PixelGAN, $16 \times 16$ PatchGAN, and $34 \times 34$ PatchGAN. The two PatchGAN networks were created following \cite{s.bjorkPotentialSequentialNonsequential2022} by modifying discriminator depth to achieve receptive field sizes of $16 \times 16$ or $34 \times 34$. Hyperparameter tuning also evaluated normalisation layers, two different weight initialisation methods, and selection of objective functions for the pretraining and fine-tuning stages.
5-fold cross-validation (CV) was used for hyperparameter tuning of the Tanzanian and Norwegian models. For Norwegian models, training and validation datasets used for CV consisted of image patches from all three Norwegian regions. Superpatches not used for test sets were divided into training and validation splits with 80\% for training and 20\% for testing in 5-fold CV. Superpatches were further divided into training (or validation) image patches of $64\times 64$ pixels with 50\% overlap allowed for training data. Data augmentation with flipping and rotation was applied to increase training data. Validation loss, based on mean and median RMSE, was used to identify optimal hyperparameters and model settings for both the Tanzanian and Norwegian models. \tabref{tab:hypbaselinerange} lists evaluated hyperparameters and search ranges for the Tanzanian and Norwegian models, where normalisation "None" refers to no normalisation layers.

\subsubsection{Summary of findings from hyperparameter tuning}

\begin{table*}[]
\caption{Selected hyperparameters and model settings for pretrained models. 
}
    \label{tab:hypbaseline}
    \centering
    \begin{tabular}{l | c c | c c}
         \hline
         \multirow{2}{*}{Hyperparameters} & \multicolumn{2}{|c}{Tanzanian models} & \multicolumn{2}{|c}{Norwegian models}\\
         & PAR U-Net  & cGAN U-NET & PAR U-Net  & cGAN U-NET \\
   \hline
   \hline 
  Batch size (BS) & 2 & 2 & 8 & 8 \\
  $\beta1$ (Adam) & 0.7 &  0.7 & 0.8 & 0.8 \\
  Learning rate (lr) & 0.0001 & 0.001 & 0.0001 & 0.001 \\
  Encoder Network &  ResNet34 &  ResNet34 & ResNet34 & ResNet34 \\
  Encoder/decoder depth  & 5 & 5 & 4 & 4 \\
  Discriminator network & --- & PixelGAN & --- & PatchGAN(16) \\
  Normalisation & None  & Instance & Instance & Batch \\ 
  $\alpha$ & 0 & 0.01 & 0 & 0.001 \\
  True target weight $\delta$ & 500 & 300 & 200 & 400\\
  Epochs & 150 & 150 & 200 & 200 \\
    \hline 
    \end{tabular}
\end{table*} 

\begin{table*}[]
\caption{Selected hyperparameters and model settings for fine-tuned models. 
}
    \label{tab:hypfine}
    \centering
        \resizebox{\textwidth}{!}{%
    \begin{tabular}{l|c|c|c|c||c|c|c|c}
        \hline
        & \multicolumn{4}{c||}{Fine-tuned Tanzanian models} & \multicolumn{4}{c}{Fine-tuned pan-regional Norwegian models} \\
        \hline
        \backslashbox{Hyperpar.}{\makecell{Pretraining:\\Fine-tuning:}} & \makecell{PAR U-Net \\\lcgan} & \makecell{PAR U-Net\\\L + \!\lfft} & \makecell{PAR U-Net\\\!\lcgan + \!\lfft\!} & \makecell{cGAN U-Net\\\!\lcgan + \!\lfft\!} & \makecell{PAR U-Net\\\lcgan} & \makecell{PAR U-Net\\\L + \!\lfft} & \makecell{PAR U-Net\\\!\lcgan + \!\lfft\!} & \makecell{cGAN U-Net\\\!\lcgan + \!\lfft\!} \\ 
   \hline 
  Batch size (BS) & 2 & 2 & 2 & 2 & 8 & 8 & 8 & 8 \\
  $\beta1$  (Adam) & 0.7 & 0.7 & 0.7 & 0.7 & 0.8 & 0.8 & 0.8 & 0.8 \\
  Learning rate (lr) & 0.0001 & 0.0001 & 0.0001 & 0.001 & 0.0001 & 0.0001 & 0.0001 & 0.001 \\
  Encoder Network & ResNet34 & ResNet34 & ResNet34 & ResNet34 & ResNet34 & ResNet34 & ResNet34 & ResNet34 \\
  Encoder/decoder depth  & 5 & 5 & 5 & 5 & 4 & 4 & 4 & 4 \\
  Discriminator network & PixelGAN  & --- & PixelGAN   &  PixelGAN & PatchGAN(16)  & --- & PatchGAN(16)   & PatchGAN(16) \\
  Normalisation & Instance & None & Instance & Instance & Instance & Instance & Instance & Batch \\
    $\alpha$ & 0.01 & 0 & 0.01 & 0.01 & 0.01 & 0 & 0.01 & 0.01 \\
    $\gamma$ & 0 & 9e-8 & 1e-7 & 1e-7 & 0 & 9e-8 & 1e-7 & 1e-7 \\
True target weight $\delta$ & 400 & 200 & 500 & 400 & 700 & 700 & 200 & 700 \\
  Epochs & 150+250 & 150+100 & 150+50 & 150+100 & 300+100 & 300+100 & 200+150 & 300+50 \\
    \hline 
    \end{tabular}}
\end{table*} 

We observed that the three ResNet networks used as convolutional encoder had similar performance, but ResNet34 was the most accurate and was therefore selected. We found that cGAN-based models optimised on the \lcgan\ objective were more accurate than those optimised on the VGAN objective, and the \lcgan\ loss was therefore selected. The evaluation of different \D\ networks showed that the Tanzanian adversarial models preferred the PixelGAN, while the $16 \times 16$ PatchGAN was preferred by the Norwegian adversarial models.

We also investigated the impact of the network's normalisation layers on the model performance. Previous work has argued that using BN in the network might harm the inherent range flexibility of the features \cite{limEnhancedDeepResidual2017, nahDeepMultiscaleConvolutional2017,  ESRGAN}. They suggest to remove the BN layers from the model architecture to increase performance and reduce computational complexity for reconstruction tasks that optimise e.g.\ the RMSE.  Motivated by \cite{limEnhancedDeepResidual2017, nahDeepMultiscaleConvolutional2017,  ESRGAN}, 
we compared batch normalisation (BN) layers to instance normalisation (IN) layers and no normalisation. Our experiments did not confirm that normalisation, and BN in particular, should be avoided. On the contrary, most models preferred BN or IN.

The potential of transfer learning was investigated by initialising the Tanzanian or Norwegian networks with or without ImageNet weights. Use of pretrained ImageNet weights requires that the input image patches from \sen must be scaled to the range [0, 1] and normalised with ImageNet mean and standard deviation. This implies that the \sen data no longer are in dB form. However, experiments showed that randomly initialised weights gave better performance than pretrained ImageNet weights. This confirms the claims from \cite{s.bjorkPotentialSequentialNonsequential2022} that avoiding normalisation and keeping the input data on dB form resulted in improved prediction of ALS-derived AGB maps. Thus, no models in this study employ pretrained ImageNet weights.

In \cite{isola}, the regularisation weight $\alpha$ was applied on the \L\ part of the generator loss function and evaluated for $\alpha\in[0,100]$. As explained in \Sec{sec:cgan} and shown in Eq.\ \eqref{eq:lsganl1}, we apply $\alpha$ on \lsgan and combine it with \L, to form \lcgan. We evaluated $\alpha$ = [0.01, 0.1, 1]. 
In accordance with \cite{isola}, we found that models trained with $\alpha = 0.01$ performed best. 

Initial experiments showed that the \L\ loss magnitude varies around $3 \times10^1$,  while the \lgan\ loss approximates $1 \times 10^0$. In contrast, the \lfft\ loss assumes magnitudes around $3 \times 10^8$. To balance the impact of the \lfft\ loss with the other loss functions, we evaluated the following range for its regularisation weight: $\gamma$ = [1e-8, 3e-8, 5e-8, 7e-8, 9e-8, 1e-7]. Our experiments show that  $\gamma$ = 9e-8 or 1e-7 is best. 

The true target weight $\delta$, used in Eq.\ \eqref{eq:masked-loss}, must be large to compensate for the strong imbalance between the numbers of true targets and pseudo-targets. We experimented with $\delta\!=\![100, 200, 300, 400, 500]$ for the Tanzanian models and $\delta\!=\![200, 300, 400, 500, 600, 700]$ for the Norwegian models. 

The selected hyperparameters for the models pretrained on the Tanzania dataset and the Norway dataset are shown in \tabref{tab:hypbaseline}. The resulting hyperparameters for the fine-tuned Tanzania models and the fine-tunes pan-regional Norwegian models are shown in \tabref{tab:hypfine}.
The hyperparameters selected for the regional Norwegian models are similar to those of their pan-regional counterparts. The exceptions are the regional PAR U-Net baseline model, which was trained for 250 epochs instead of 200, and the regional model pretrained on the \L\ loss and fine-tuned on the \lcgan\ loss, which was fine-tuned for 250 epochs instead of 100. The tables report the number of epochs as $E_{p}+E_{f}$, denoting $E_{p}$ epochs of pretraining and $E_{f}$ in fine-tuning.

\end{document}